\documentclass[10pt, conference, compsocconf]{IEEEtran}
\ifCLASSINFOpdf
\else
\fi
\newcommand{\ignore}[1]{}
\usepackage[pass]{geometry}
\usepackage{fancyhdr}
\usepackage[normalem]{ulem}
\usepackage[hyphens]{url}
\usepackage{color}
\usepackage{soul}
\usepackage{longtable}
\usepackage[normalem]{ulem}
\usepackage[hyphens]{url}
\usepackage{amsmath}

\usepackage[boxed,ruled]{algorithm2e}
\usepackage[numbers,sort&compress]{natbib}
\usepackage{makecell}
\usepackage[para,online,flushleft]{threeparttable}
\usepackage{amssymb}
\usepackage{color}
\usepackage{amsmath}
\usepackage{caption}
\usepackage{graphicx}
\usepackage{placeins}
\usepackage{float}
\usepackage{subfigure}
\usepackage[normalem]{ulem}
\usepackage[hyphens]{url}
\usepackage{multicol}
\usepackage{multirow}
\usepackage{fancyhdr}
\usepackage{mathtools}

\DeclarePairedDelimiter\floor{\lfloor}{\rfloor}
\usepackage{xcolor}
\usepackage{soul}

\usepackage{color, soul}
\soulregister\cite7
\soulregister\ref7
\soulregister\underline7

\usepackage[nodisplayskipstretch]{setspace}
\renewcommand\baselinestretch{0.92}
\PassOptionsToPackage{bookmarks={false}}{hyperref}
\begin{document}
%

% paper title
% can use linebreaks \\ within to get better formatting as desired
\title{ \vskip -2.5em E-RNN: Design Optimization for Efficient Recurrent Neural Networks in FPGAs  \vskip -1.5em}

% author names and affiliations
% use a multiple column layout for up to two different
% affiliations

\author{
\IEEEauthorblockN{Zhe Li$^{1*}$, Caiwen Ding$^{2*}$, Siyue Wang$^2$, Wujie Wen$^3$, Youwei Zhuo$^4$, Chang Liu$^5$, \\Qinru Qiu$^1$, Wenyao Xu$^6$, Xue Lin$^2$, Xuehai Qian$^4$, and Yanzhi Wang$^2$\\$^*$These authors contributed equally.
}
\IEEEauthorblockA{$^1$Syracuse University, $^2$Northeastern University,  $^3$Florida International University,  \\ $^4$University of Southern California, $^5$Carnegie Mellon University, $^6$ SUNY University at Buffalo.\\
$^1$\{zli89, qiqiu\}@syr.edu,  $^2$\{ding.ca, wang.siy\}@husky.neu.edu,   $^2$\{xue.lin, yanz.wang\}@northeastern.edu, \\ $^3$wwen@fiu.edu, $^4$\{youweizh,xuehai.qian\}@usc.edu,
$^5$cliu5@andrew.cmu.edu,$^6$wenyaoxu@buffalo.edu.
% \vskip -1.2em 
}
}
% \and
% \IEEEauthorblockN{Authors Name/s per 2nd Affiliation (Author)}
% \IEEEauthorblockA{line 1 (of Affiliation): dept. name of organization\\
% line 2: name of organization, acronyms acceptable\\
% line 3: City, Country\\
% line 4: Email: name@xyz.com}

\maketitle

\begin{abstract}
Recurrent Neural Networks (RNNs) are becoming increasingly important for time series-related applications which require efficient and real-time implementations. The two major types are Long Short-Term Memory (LSTM) and Gated Recurrent Unit (GRU) networks. It is a challenging task to have real-time, efficient, and accurate hardware RNN implementations because of the high sensitivity to imprecision accumulation and the requirement of special activation function implementations.
Recently two works have focused on FPGA implementation of inference phase of LSTM RNNs with model compression. First, ESE uses a weight pruning based compressed RNN model but suffers from irregular network structure after pruning. The second work C-LSTM mitigates the irregular network limitation by incorporating block-circulant matrices for weight matrix representation in RNNs, thereby achieving simultaneous model compression and acceleration.

A key limitation of the prior works is the lack of a systematic design optimization framework of RNN model and hardware implementations, especially when the block size (or compression ratio) should be jointly optimized with RNN type, layer size, etc. In this paper, we adopt the block-circulant matrix-based framework, and present the Efficient RNN (E-RNN) framework for FPGA implementations of the Automatic Speech Recognition (ASR) application. The overall goal is to improve performance/energy efficiency under accuracy requirement. We use the alternating direction method of multipliers (ADMM) technique for more accurate block-circulant training, and present two design explorations providing guidance on block size and reducing RNN training trials.  Based on the two observations, we decompose E-RNN in two phases: Phase I on determining RNN model to reduce computation and storage subject to accuracy requirement, and Phase II on hardware implementations given RNN
model, including processing element design/optimization, quantization, activation implementation, etc. \footnote{The code is available in \url{https://github.com/lz1313/BlockCIrculantRNN}.}
 Experimental results on actual FPGA deployments show that E-RNN achieves a maximum energy efficiency improvement of 37.4$\times$ compared with ESE, and more than 2$\times$ compared with C-LSTM, under the same accuracy.   

\end{abstract}


% For peer review papers, you can put extra information on the cover
% page as needed:
% \ifCLASSOPTIONpeerreview
% \begin{center} \bfseries EDICS Category: 3-BBND \end{center}
% \fi
%
% For peerreview papers, this IEEEtran command inserts a page break and
% creates the second title. It will be ignored for other modes.
\IEEEpeerreviewmaketitle

\section{Introduction}

Recurrent Neural Networks (RNNs) represent an important class of machine learning techniques that are specialized for processing sequential data \cite{Goodfellow-et-al-2016}. RNNs have wide applications in speech recognition, natural language processing, scene and semantic understanding, time series analysis, etc. Many of these applications require efficient and real-time implementations. The two major types of RNNs with the broadest applications and highest performance are the \emph{Long Short-Term Memory} (LSTM) unit \cite{hochreiter1997long} and the \emph{Gated Recurrent unit} (GRU) \cite{cho2014properties}. LSTM and GRU RNNs are computationally intensive but can effectively overcome \emph{vanishing} and \emph{exploding} gradient problems \cite{pascanu2013difficulty} of traditional RNNs.

As RNNs are related to time series analysis and used for making temporal decisions, the real-time, high-efficiency hardware implementations of RNNs are becoming imperative. Recently, there have been extensive investigations in industry and academia \cite{alwani2016fused,shen2017maximizing,chen2014diannao,tpu,ren2017sc,merolla2014million,sharma2016high,shen2017escher,ovtcharov2015toward,ovtcharov2015accelerating,sharma2016dnnweaver} on hardware acceleration of (the inference phase of) feedforward \emph{Deep Neural Networks} (DNNs)\footnote{We differentiate between feedforward DNNs used mainly for image classification and cycle-based RNNs used mainly for sequential data processing.}, in both FPGA and ASIC accelerations. Model compression and algorithm-level acceleration of DNNs have also been investigated, including weight quantization \cite{lin2016fixed,wu2016mobile}, connection pruning \cite{han2015deep,han2015learning}, and low rank approximation \cite{jaderberg2014speeding,tai2015convolutional}. Despite all this effort, there have been limited contributions in prior work on the subject of efficient RNN implementations, at least the inference phase which requires real-time performance in power-budgeted systems.
In fact, hardware implementations and model compression of RNNs exhibit unique challenges. First, RNNs are very sensitive to accumulation of imprecisions, due to both model compression and bit quantization. Additionally, for LSTM/GRU RNNs, there are special operations like point-wise multiplications and special activation functions like tanh (hyperbolic tangent) \cite{hochreiter1997long,cho2014properties,sak2014long}, which require accurate and efficient hardware implementations. 

As a representative work on implementing LSTMs on FPGAs, the ESE \cite{han2017ese} implements the inference phase of sparse LSTM model obtained by the parameter pruning method \cite{han2015deep,han2015learning}. The ESE achieves higher energy efficiency than GPU, but its performance is lower. This is due to (i) the limited compression ratio for LSTMs (4-6$\times$ when indices are accounted for), (ii) the irregular network structure after pruning, and (iii) the inefficient implementation of activations and indices.

\begin{figure}[t]
    \centering
    \vspace{-0.5em}
    \includegraphics[width=0.7\columnwidth]{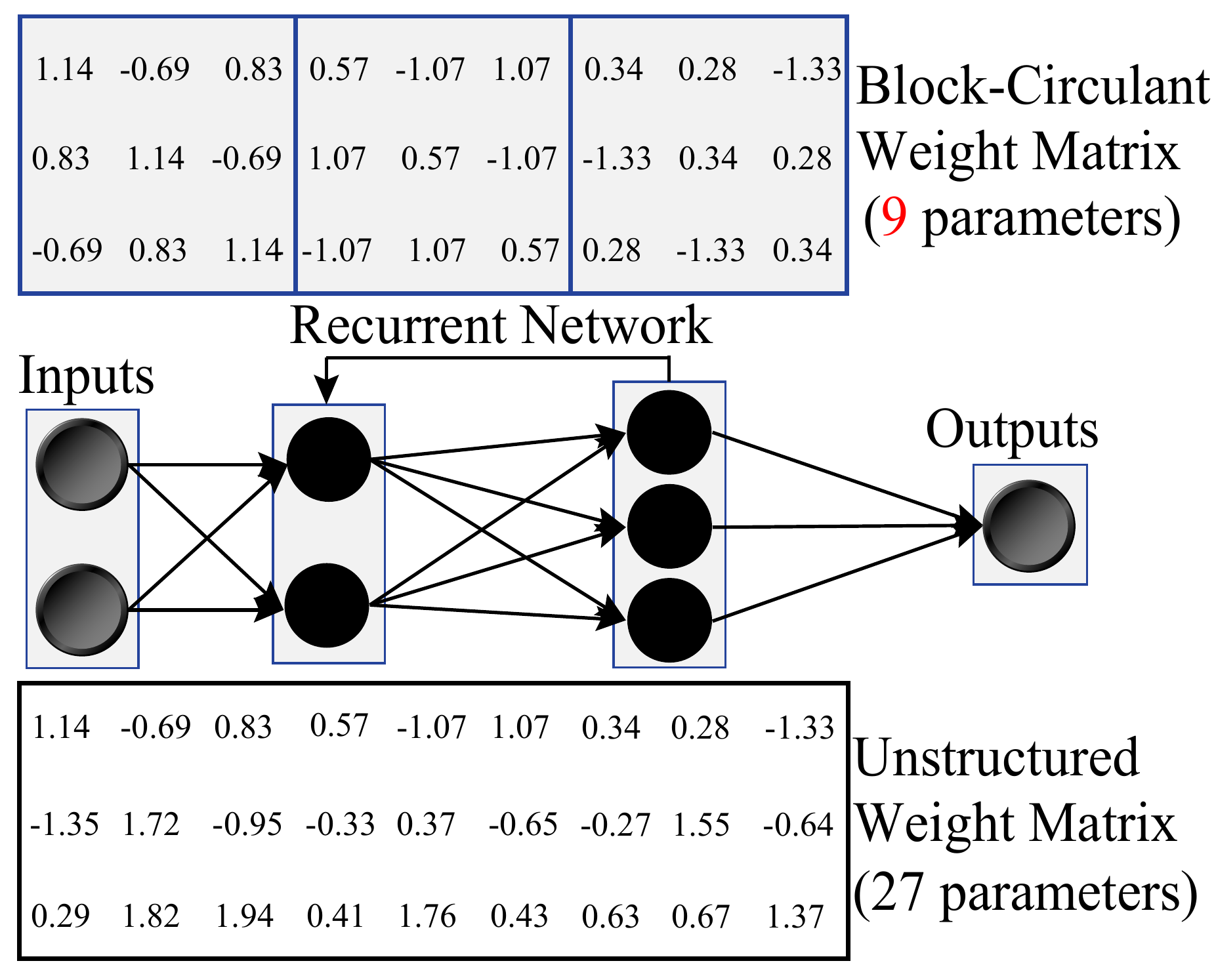} 
    \vskip -0.5em
    \caption{An illustrative example of block-circulant matrices for RNN weight representation.}
    \label{fig:Intro_Block_Matrix}
 \end{figure}

In order to exploit the full computing power of FPGAs and overcome the irregularity issue, the recent work C-LSTM \cite{Wang2018clstm} has adopted block-circulant matrices \cite{ding2017c,liao2017energy} for weight matrix representations in LSTM RNNs, thereby achieving simultaneous model compression and acceleration. Fig.~\ref{fig:Intro_Block_Matrix} shows an illustrative example. 
A block-circulant matrix consists of a set of square circulant submatrices (blocks). 
In a circulant matrix, each row (or column) vector is a circulant reformat of the other row (column) vectors.
Therefore, each submatrix can be represented by a vector. The first obvious benefit is storage size reduction from O($n^2$) to O($n$). 
In LSTM RNN, the major computation is $\textbf{Wx}$ of weight matrix $\textbf{W}$ and vector $\textbf{x}$, where $\textbf{W}$ is now block-circulant. The \emph{Fast Fourier Transform} (FFT) method could be utilized for acceleration, and the {computational complexity is reduced from O($n^2$) to O($n\log n$)}. 
In addition to the computational and storage complexity reductions, the block-circulant matrix-based compression generates regular, structured weight matrices, which is amenable to efficient hardware accelerations.

Overall speaking, the block-circulant matrix-based framework allows us to achieve a fine-grained trade-off between \emph{accuracy} and \emph{compression/acceleration ratio}.
A larger block size should be selected to achieve a higher compression ratio, however, it may degrade the accuracy.
The smaller block sizes provide higher accuracy, but less 
compression ratio. 

The prior work focus on the efficient implementation of RNN inference phase given a pre-computed RNN model. They did not provide a systematic method to perform design optimization. When the block size (or degree of model compression) needs to be optimized together with the network type/size and different block sizes can be utilized for different parts of a network, a significant increase in the number of RNN training trials will be needed for design optimization. 
Moreover, the design optimization needs to be judiciously performed based on the overall accuracy and performance requirements, as well as the computation and storage resources of the hardware platform (e.g., FPGA). An algorithm-hardware crosslayer framework is desirable.

In this work, we focus on block-circulant matrix-based RNN implementations and aim to mitigate these limitations. We propose {fast and effective design optimizations for RNN implementation}, in which the term \emph{fast} refers to reducing the number of RNN training trials to arrive at a close-to-optimal solution, and \emph{effectiveness} is defined in terms of performance and energy efficiency in (FPGA) hardware implementation under overall accuracy requirements. The target application is Automatic Speech Recognition (ASR), which is a representative and computation-intensive application of (LSTM and GRU) RNNs and is also the focus of \cite{han2017ese}. Different from prior works, we applied ADMM~\cite{boyd2011alternating} to train the block circulant based RNN models to achieve better accuracy. ADMM is a powerful method for solving non-convex optimization problems with combinatorial constraints. 
% We have released the related codes online as follows:
% \href{https://github.com/AnonymousAccount1313/BlockCIrculantRNN}{\underline{https://github.com/AnonymousAccount1313/BlockCirculant}\\\underline{RNN.}}
% {\underline{https://github.com/lz1313/BlockCIrculantRNN}}

To provide some high-level guidelines, we first perform {two design explorations on the RNN model}: The first one is top-down from the algorithm level, and clearly demonstrates that block size optimization should be prioritized over layer size optimization under the overall accuracy constraint. The second one is a bottom-up framework focusing on computation reductions, and effectively sets a proper range of block size optimization. These two observations can effectively reduce the number of training trials in design optimization.

 \begin{figure} [t]
 	\centering
 	\vspace{-1.2em}
 	\includegraphics[width=0.65\columnwidth]{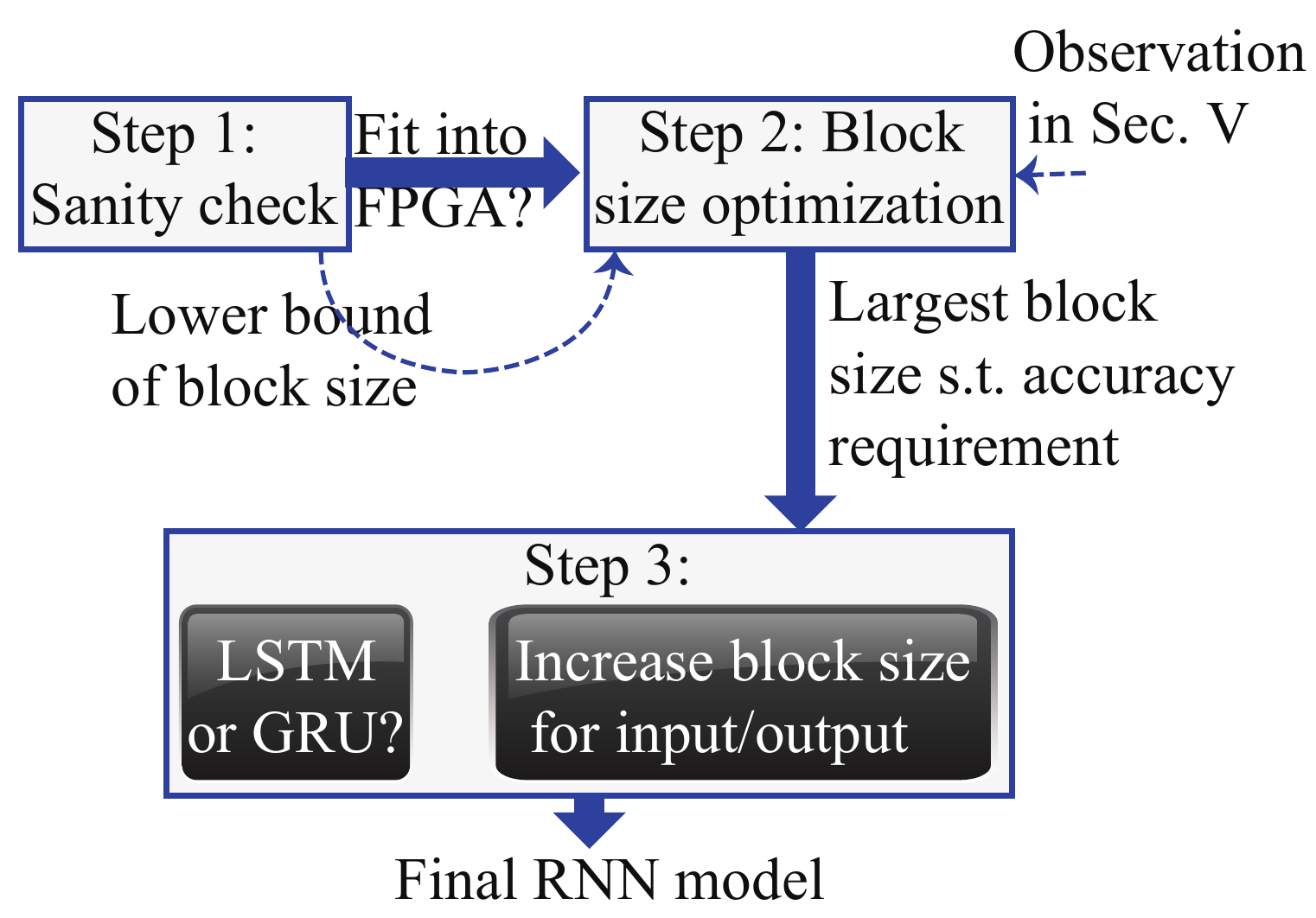}	
 	\vskip -0.5em
 	\caption{The Phase-I algorithm of E-RNN.}
 	\label{fig:phase1}
 \end{figure}

Based on these two observations, we propose the {E-RNN design optimization framework of RNN implementation on FPGAs}. The proposed framework is also applicable to ASICs. The optimization objectives are performance and energy efficiency under the overall accuracy requirement. The optimization variables include model type (LSTM, GRU, etc.) selection, block size and layer size optimization, hardware implementation structure and parallelism degree, quantization and activation functions, etc. We divide the overall design optimization into two phases. \emph{Phase I} lies at the interface between algorithm and hardware and determines RNN model specifications, including model type, layer size, and block size, under the overall accuracy constraint as shown in Fig.~\ref{fig:phase1}. The number of training trials is effectively reduced by leveraging the above observations. The RNN model can be fully accommodated using on-chip BRAM of FPGA through this phase. 
\emph{Phase II} focuses on hardware-oriented optimization given the RNN model, and determines the hardware implementation structure, the number of \emph{processing elements} (PEs), quantization scheme and activation function implementations, etc.
% Our network design optimization approach is useful for other NN designs targetted at FPGAs as well. 
% The major contributions of this work are: 
We conclude the contribution of E-RNN in two-fold:
(i) At software level, we use ADMM-based training for deriving block-circulant matrix-based RNN representation.
ADMM-based training is compatible with recent progress in stochastic gradient descent (e.g., ADAM), which is not supported in the training method of C-LSTM \cite{Wang2018clstm}. ADMM-based training provides an effective means to deal with the structure requirement in weight matrices, thereby enhancing accuracy and training speed.
%It generalizes well for other training optimization method due to its universality; 
(ii) At hardware level, we propose a systematic design framework and hardware optimization using HLS, to achieve alternative designs (LSTM vs. GRU) for RNNs, and to limit the design range and accelerate the design exploration. The systematic framework also works for other DNN designs targeted at FPGAs due to the regularity of block-circulant matrix. 
% The very good hardware testing results with negligible accuracy degradation shows the performance improvement through the systematic design optimization technique. 
% (i) training of block-circulant model using ADMM; (ii) alternative designs (LSTM vs. GRU) for RNNs; and (iii) systematic design exploration and hardware architecture.
% Specifically, we show: how to achieve a faster and systematic design exploration targeting hardware HLS when incorporating model compression technique. This is lacking in the previous work.
Experimental results on actual FPGA deployments shows that the proposed E-RNN framework achieves a significant energy efficiency improvement of 37.4$\times$ compared with ESE \cite{han2017ese} under the same accuracy degradation, and energy efficiency improvement of over 2$\times$ compared with C-LSTM~\cite{Wang2018clstm}. 

\section{Background on RNN Cells}

\subsection{Long short-term memory (LSTM)}
\begin{figure} [tb]
 	\centering
 	\vspace{-1.2em}
 	\includegraphics[width=0.95\columnwidth]{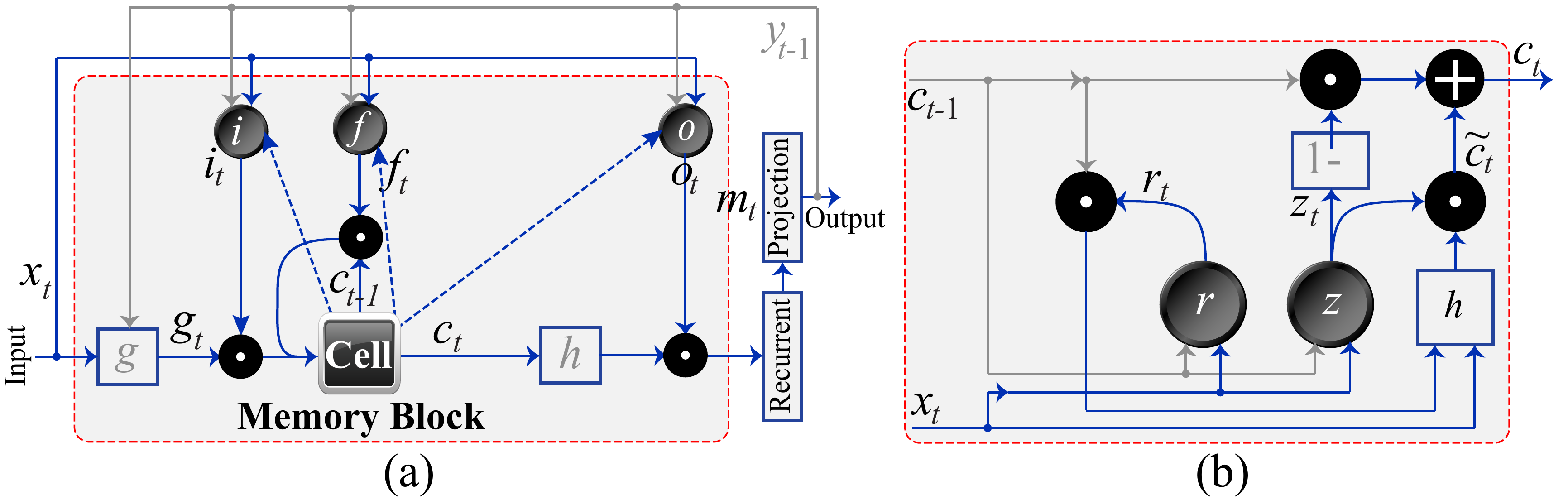}	
 	\caption{(a) An LSTM based and (b) a GRU based RNN architecture.}
 	\label{fig:LSTM_GRU}
 \end{figure}

Modern large scale Automatic Speech Recognition (ASR) systems take advantage of LSTM-based RNNs as their acoustic models. An LSTM model consists of large matrices which is the most computational intensive part among all the steps of the ASR procedure. We focus on a representative LSTM model presented in \cite{sak2014long} whose architecture is shown in Fig.~\ref{fig:LSTM_GRU} (a).
% In the LSTM-based RNN, the input at time $T$ depends on the output at $T-1$. The LSTM's block contains a special memory cell storing the temporal state of the network.
% It also contains special multiplicative units called \emph{gates}: \emph{input gate}, \emph{output gate} and \emph{forget gate}.
% The input gate $i$ controls the amount of input contribution into the memory cell. The output gate $o$ controls the output value, while the forget gate $f$ determines the scale of the previous state of the cell which can adaptively forget the cell's memory.
% Besides the basic three well-known gates and the cell state, this LSTM model also introduced \emph{peephole} \cite{gers2000recurrent} and \emph{projection layer} \cite{sak2014long} for the sake of better learning. The peephole adds the scaled previous cell state to the three gates. The scales are determined by three diagonal matrices. The projection layer linearly transforms the output to a lower dimension.
%  \begin{figure} [tb]
%  	\centering
%  	\includegraphics[width=0.45\columnwidth]{figs/Fig_GRU1.pdf}	
%  	\vskip -0.5em
%  	\caption{A GRU based RNN architecture.}
%  	\label{fig:GRU}
%  \end{figure}
An LSTM-based RNN accepts an input vector sequence $\mathbb{X}= (\mathbf{x}_1; \mathbf{x}_2; \mathbf{x}_3; ...; \mathbf{x}_T)$ (each of $\mathbf{x}_t$ is a vector corresponding to time $t$) with the output sequence from last step $\mathbb{Y}^{T-1} = (\mathbf{y}_0; \mathbf{y}_1; \mathbf{y}_2; ...; \mathbf{y}_{T-1})$ (each of $\mathbf{y}_t$ is a vector).
It computes an output sequence $\mathbb{Y} = (\mathbf{y}_1;\mathbf{y}_2; \mathbf{y}_3; ...; \mathbf{y}_T )$ by using the following equations iteratively from $t = 1$ to $T$:
\begin{subequations}\label{eqn:model}\small
\begin{align}
    \mathbf{i}_t &= \sigma(\mathbf{W}_{ix}\mathbf{x}_t +\mathbf{W}_{ir}\mathbf{y}_{t-1} + \mathbf{W}_{ic}\mathbf{c}_{t-1}+\mathbf{b}_i), \\
    \mathbf{f}_t &= \sigma(\mathbf{W}_{fx}\mathbf{x}_t +\mathbf{W}_{fr}\mathbf{y}_{t-1} + \mathbf{W}_{fc}\mathbf{c}_{t-1}+\mathbf{b}_f), \\
    \mathbf{g}_t &= \sigma(\mathbf{W}_{cx}\mathbf{x}_t + \mathbf{W}_{cr}\mathbf{y}_{t-1} + \mathbf{b}_c), \\
    \mathbf{c}_t &= \mathbf{f}_t \odot \mathbf{c}_{t-1} + \mathbf{g}_t \odot \mathbf{i}_t, \\
    \mathbf{o}_t &= \sigma(\mathbf{W}_{ox}\mathbf{x}_t + \mathbf{W}_{or}\mathbf{y}_{t-1} + \mathbf{W}_{oc}\mathbf{c}_{t}+\mathbf{b}_o), \\
    \mathbf{m}_t &= \mathbf{o}_t \odot \mathbf h(\mathbf{c}_t), \\
    \mathbf{y}_t &= \mathbf{W}_{ym}\mathbf{m}_t,
\end{align}
\end{subequations}
where symbols $\mathbf{i}$, $\mathbf{f}$, $\mathbf{o}$, $\mathbf{c}$, $\mathbf{m}$, and $\mathbf{y}$ are respectively the input gate, forget gate, output gate, cell state, cell output, and projected output~\cite{sak2014long};
the $\odot$ operation denotes the point-wise multiplication, and the $+$ operation denotes the point-wise addition.
The $\mathbf{W}$ terms denote weight matrices (e.g. $\mathbf{W}_{ix}$ is the matrix of weights from the input vector $\mathbf{x}_t$ to the input gate), and the $\mathbf{b}$ terms denote bias vectors. 
Please note $\mathbf{W}_{ic}$, $\mathbf{W}_{fc}$, and $\mathbf{W}_{oc}$ are diagonal matrices for peephole connections~\cite{gers2000recurrent}, thus they are essentially a vector. As a result, the matrix-vector multiplication like $\mathbf{W}_{ic}\mathbf{c}_{t-1}$ can be calculated by the $\odot$ operation.
$\sigma$ is the logistic activation function and $\mathbf{h}$ is a user defined activation function. Here we use hyperpolic tangent (tanh) activation function as $\mathbf{h}$.

In the above equations, we have nine matrix-vector multiplications (excluding peephole connections which can be calculated by $\odot$). 
In one gate/cell, $\mathbf{W}_{\ast x}\mathbf{x}_t + \mathbf{W}_{\ast r}\mathbf{y}_{t-1}$ can be combined in one matrix-vector multiplication by concatenating the matrix and vector as $\mathbf{W}_{\ast (xr)}[{\mathbf{x}_{t}^T, \mathbf{y}_{t-1}^T}]^T$. 
The four gate/cell matrices can be concatenated and calculated through one matrix-vector multiplication as $\textbf{W}_{(ifco) (xr)}[\textbf{x}_t^T, \textbf{y}_{t-1}^T]^T$.
Thus, we can compute the above equations with two matrix-vector multiplications, i.e. $\textbf{W}_{(ifco) (xr)} $$[\textbf{x}_t^T, \textbf{y}_{t-1}^T]^T$ and $\textbf{W}_{ym}\textbf{m}_t$.

\subsection{Gated recurrent units (GRU)}

The GRU is a variation of the LSTM as introduced in \cite{cho2014learning}. It combines the forget and input gates into a single ``update gate''. It also merges the cell state and hidden state, and makes some other changes. The architecture is shown in Fig.~\ref{fig:LSTM_GRU} (b).
Similarly, it 
%computes an output sequence $\mathbb{Y} = (\mathbf{y}_1;\mathbf{y}_2; \mathbf{y}_3; ...; \mathbf{y}_T )$ by using the 
follows equations iteratively from $t = 1$ to $T$:
\begin{subequations}\label{eqn:grumodel}\small
\begin{align}
    \mathbf{z}_t &= \sigma(\mathbf{W}_{zx}\mathbf{x}_t +\mathbf{W}_{zc}\mathbf{c}_{t-1} + \mathbf{b}_z), \\
    \mathbf{r}_t &= \sigma(\mathbf{W}_{rx}\mathbf{x}_t +\mathbf{W}_{rc}\mathbf{c}_{t-1}+ \mathbf{b}_r), \\
    \mathbf{\tilde{c}}_t &= \mathbf h(\mathbf{W}_{\tilde{c}x}\mathbf{x}_t + \mathbf{W}_{\tilde{c}c}(\mathbf{r}_t \odot  \mathbf{c}_{t-1}) + \mathbf{b}_{\tilde{c}}),\\
    \mathbf{c}_t &= (1 - \mathbf{z}_t) \odot \mathbf{c}_{t-1}+ \mathbf{z}_t \odot \mathbf{\tilde{c}}_t
\end{align}
\end{subequations}
where symbols $\mathbf{z}$, $\mathbf{r}$, $\mathbf{\tilde{c}}$, $\mathbf{c}$ are respectively the update gate, reset gate, reset state, and cell state;
the $\odot$ operation denotes the point-wise multiplication, and the $+$ operation denotes the point-wise addition.
The $\mathbf{W}$ terms denote weight matrices (e.g. $\mathbf{W}_{zx}$ is the matrix of weights from the input vector $\mathbf{x}_t$ to the reset gate).
$\sigma$ is the logistic activation function and $\mathbf{h}$ is a user defined activation function. Here we use tanh activation function as $\mathbf{h}$.
Note that a GRU has two gates (update and reset), while an LSTM has three gates (input, forget, output).
GRUs do not have the output gate that is present in LSTMs. Instead, the cell state is taken as the output.
The input and forget gates are coupled by an update gate $\mathbf{z}$, and the reset gate $\mathbf{r}$ is applied directly to the previous cell state. 

In the above set of equations, we have six matrix-vector multiplications. 
In the reset and update gates, $\mathbf{W}_{\ast x}\mathbf{x}_t + \mathbf{W}_{\ast c}\mathbf{c}_{t-1}$ can be combined/fused in one matrix-vector multiplication by concatenating the matrix and vector as $\mathbf{W}_{\ast (xc)}[{\mathbf{x}_{t}^T, \mathbf{c}_{t-1}^T}]^T$. 
Furthermore, the reset and update gate matrices can also be concatenated and calculated through one matrix-vector multiplication as $\textbf{W}_{(rz) (xc)}[\mathbf{x}_t^T, \textbf{c}_{t-1}^T]^T$.
In this way, we compute the above equations with three matrix-vector multiplications, i.e. $\mathbf{W}_{(rz) (xc)}[\mathbf{x}_t^T, \mathbf{c}_{t-1}^T]^T$, $\mathbf{W}_{\tilde{c}x}\mathbf{x}_t$, and $\mathbf{W}_{\tilde{c}c}(\mathbf{r}_t \odot  \mathbf{c}_{t-1})$.

\section{Block-Circulant Matrices for RNN Models}

%Reference \cite{Wang2018clstm} has incorporated block-circulant matrices~\cite{cheng2015exploration,ding2017circnn} for weight representation in LSTM models. 
% \hl{Suppose we partition an arbitrary weight matrix with size $m \times n$ of an RNN layer into $p\times q$ blocks, where each block is a squared submatrix with size $L_b\times L_b$ and $p = m/L_b$, $q = n/L_b$. The block-circulant matrix-based RNN representation uses a \emph{circulant} matrix for each block (or squared submatrix). In  a  circulant matrix, each row (or column) vector is a circulant reformat of the other row (column) vectors.  Therefore, each submatrix with size $L_b\times L_b$  can  be  represented  by  a  vector of size $L_b$. The number of parameters in a weight matrix is reduced from $p\cdot q\cdot L_b\cdot L_b$ by a full matrix to $p\cdot q\cdot L_b$ by a block-circulant matrix.  
% (If $m$ or $n$ cannot be divided by $L_b$ completely with the remainder of $r_m$ and $r_n$, respectively, we pad 0's to make the whole matrix to size $(m+L_b-r_m)\times(n+L_b-r_n)$.)}

Overall, it is possible to {\em simultaneously} achieve significant reductions in {\em both computational and storage} complexity, for {\em both inference and training}. This is especially crucial for hardware implementations.

We are not forcing the block-circulant format onto a trained RNN weight matrix. Indeed, the ADMM training to be discussed in Sec. \ref{sec:ADMM} will directly result in RNN weight matrices in the block-circulant format.
From the perspective of matrix theory, the block-circulant matrices has shown the same ``effectiveness'' as the full matrices in representing RNNs as discussed in~\cite{zhao2017theoretical}. 
In practice, the block size represents a trade-off between accuracy and storage/computation complexity. There is an upper bound on the block size with minor accuracy loss.

%\hl{The block circulant matrices are trained from scratch \cite{Wang2018clstm} using stochastic gradient descent (SGD) or from the pre-trained models of full matrices (in this work) using SGD and Alternating direction method of multipliers (ADMM).
%No error is accumulated through layers as the weight matrices are trained in block-circulant format instead of being simply converted.
%The accuracy degradation compared with normal matrices is typically minor.}

% In this section, we provide a discussion of the inference algorithm for block-circulant matrix-based RNNs. We use the LSTM model as an illustrative example, but the framework is applicable to GRU as well. After that we present the limitation of \cite{Wang2018clstm}, which serves as the motivation of proposed work on design optimizations.

\subsection{Block-Circulant Matrices-Based Inference}
The primary idea of block-circulant matrix-based LSTM is to represent the original arbitrary weight matrix $\textbf{W}\in \mathbb{R}^{m\times n}$ with an array of equal-size square sub-matrices (i.e., \emph{blocks}), where each sub-matrix is a \emph{circulant} matrix.
Assume there are $p \times q$ blocks after partitioning the matrix $\mathbf{W}$, where $p = \frac{m}{L_b}$ and $q=\frac{n}{L_b}$. Here $L_b$ is the \emph{block size}. Then $\mathbf{W} = [\mathbf{W}_{ij}]$, $i \in \{1 \dots p\}$, $j \in \{1 \dots q\}$. 

\begin{figure}[t]
 	\centering
 	\vspace{-1.1em}
 	\includegraphics[width=0.8\columnwidth]{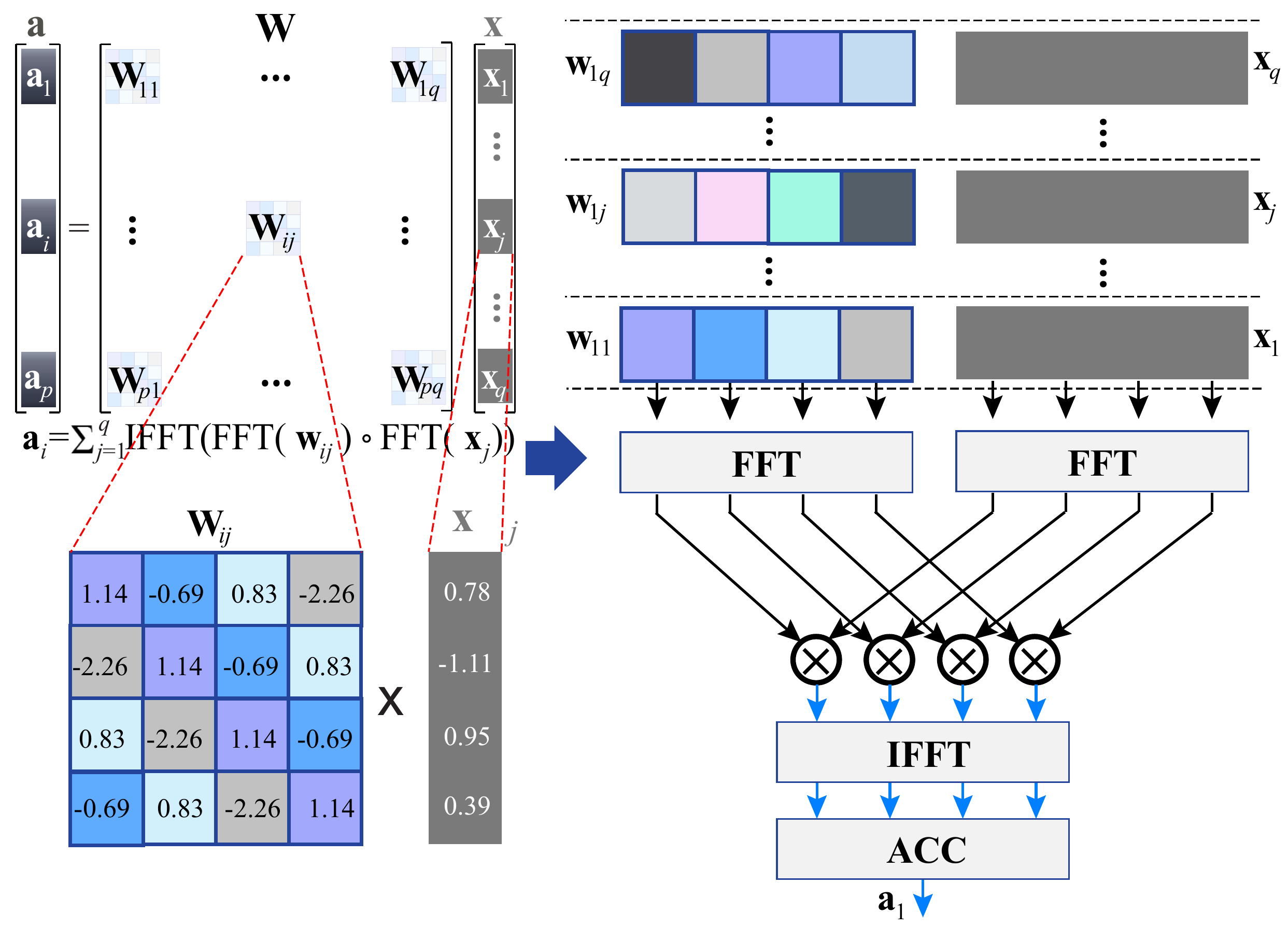}	
 	\vskip -0.5em
 	\caption{An illustration of FFT-based calculation in block-circulant matrix multiplication.}
 	\label{fig:block_matrix}
 \end{figure}

Each circulant matrix $\mathbf{W}_{ij}$ can be defined by a vector $\mathbf{w}_{ij}$. More specifically, $\mathbf{w}_{ij}$ is the first row vector of $\mathbf{W}_{ij}$; the second row vector of $\mathbf{W}_{ij}$ is a circulation of the first row vector, and so on. Fig. \ref{fig:block_matrix} provides an example of circulant matrix $\mathbf{W}_{ij}$. The storage complexity of a block-circulant weight matrix is significantly reduced since we only need to store one vector $\mathbf{w}_{ij}$ for each circulant matrix $\mathbf{W}_{ij}$.
As a result, we have the ability to store all the weights matrices (i.e.,
% ${W}_{ix}$,  ${W}_{ir}$, ${W}_{fx}$, ${W}_{fr}$, ${W}_{cx}$, ${W}_{cr}$, ${W}_{ox}$, and ${W}_{or}$
${\mathbf{W}}_{*(xr)}$) and the projection matrix ${\mathbf{W}}_{ym}$ in block RAM (BRAM), thereby significantly improving the FPGA performance.  Additionally, the input feature $\mathbf{x}$, bias $\mathbf{b}$ ($\mathbf{b}_{i}$, $\mathbf{b}_{f}$, and $\mathbf{b}_{o}$), and diagonal matrices $\mathbf{W}_{c}$ ($\mathbf{W}_{ic}$, $\mathbf{W}_{fc}$, and $\mathbf{W}_{oc}$) can also be stored in BRAM due to a small quantity of corresponding parameters.

 Since a weight matrix $\mathbf{W}$ is now partitioned into $p\times q$ blocks, correspondingly, the input $\mathbf{x}$ is also partitioned as $\mathbf{x} = [\mathbf{x}^T_1, \mathbf{x}^T_2, \dots, \mathbf{x}^T_q]^T$, $\mathbf{x}_j\in\mathbb{R}^{L_b}$.
Then, the \emph{forward propagation} process in the inference phase is given by (with bias and activation function omitted):
% \vspace{-0.1in}
\begin{equation}\label{eqn:matrix-vector}\small
\mathbf{a}
=
\mathbf{Wx} 
=
\begin{bmatrix}
         \sum_{j=1}^q \mathbf{W}_{1j} \mathbf{x}_j   \\
         \sum_{j=1}^q \mathbf{W}_{2j} \mathbf{x}_j   \\
         \dots \\
         \sum_{j=1}^q \mathbf{W}_{pj} \mathbf{x}_j  
\end{bmatrix}
=
\begin{bmatrix}
         \mathbf{a}_1   \\
         \mathbf{a}_2   \\
         \dots \\
         \mathbf{a}_p
\end{bmatrix},
\end{equation}
where $\mathbf{a}_i \in \mathbb{R}^{L_b}$ is a column vector.
We can see the calculation of $\mathbf{Wx}$ is reduced to the calculation of $\mathbf{W}_{ij} \mathbf{x}_j$'s.
Then according to the \emph{circulant convolution theorem} \cite{pan2012structured,bini1996polynomial}, the calculation of $\mathbf{W}_{ij} \mathbf{x}_j$ can be performed as
% \vspace{-0.1in}
\begin{equation}\label{eqn:FFT}\small
\mathbf{W}_{ij} \mathbf{x}_j=\text{IFFT}\big(\text{FFT}(\mathbf{w}_{ij})\odot\text{FFT}(\mathbf{x}_j)\big),
\end{equation}
where $\odot$ denotes element-wise multiplications, and FFT and IFFT denote Fast Fourier Transform (FFT) and inverse FFT, respectively.
The computational complexity of $\mathbf{Wx}$ is reduced from $O(n^2)$ by direct matrix-vector multiplication to $O(pqL_b\log L_b)$ by the ``FFT$\rightarrow$element-wise multiplication$\rightarrow$IFFT'' procedure in Eqn. (\ref{eqn:FFT}), which is equivalent to $O(n\log n)$ for small $p$, $q$ values. As a result, the simultaneous acceleration and model compression compared with the original LSTM can be achieved for the inference process.

The \emph{backward propagation} process in the training phase can also be implemented using block-circulant matrices, which is similar to the procedure in~\cite{ding2017circnn}. It is important to understand that during training, the block-circulant matrix-based approach {directly trains weight matrices in the block-circulant format by training only one vector for each block (i.e., circulant matrix)}.

\subsection{ADMM-Based Training}
\label{sec:ADMM}

% \hl{\textbf{Alternating direction method of multipliers \\ (ADMM) based block circulant matrix training.}}
% \hl{
% The optimization method used in CirCNN~\cite{ding2017circnn} can not incorprate with Adam optimization. On the other hand, directly training the original weight matrices in the block ciuculant format will incur a large number of equality constraints (to maintain the structure). This makes the training problem inefficient to solve using the conventional SGD. Therefore, we utilize ADMM to efficiently solve the optimization problem, and a large number of equality constraints can be avoided. ADMM generalizes well for other training optimization methods due to its universality.}

Consider an optimization problem $\min_{{\bf{x}}} f(\bf{x})$ with combinatorial constraints. This problem is difficult to solve directly using optimization tools \cite{zhang2018systematic}. Through the application of ADMM~\cite{boyd2011distributed,jin2017deep}, the original optimization problem is decomposed into two subproblems, and will be iteratively solved until convergence. The first subproblem is $\min_{{\bf{x}}} f({\bf{x}})+q_1(\bf{x})$ where $q_1(\bf{x})$ is a differentiable, quadratic term. This subproblem does not have combinatorial constraints and can be solved using traditional optimization method, e.g., SGD for RNN training. The second subproblem is $\min_{{\bf{x}}} g({\bf{x}})+q_2(\bf{x})$, where $g({\bf{x}})$ corresponds to the original combinatorial constraints and $q_2(\bf{x})$ is also quadratic. For special types of combinatorial constraints, including structured matrices, quantization, etc., the second subproblem can be optimally and analytically solved, as shown in the following discussions.

Consider an RNN model with $N$ layers. The collection of weights in layer $l$ is denoted by ${\bf{W}}_{l}$. The loss function is denoted by $f \big( \{{\bf{W}}_{l}\}_{l=1}^N\big)$. 
Let $({\bf{W}}_{l})_{ij}$ with dimension $L_b\times L_b$ denote the $ij^{th}$ block in the structured matrix that ${\bf{W}}_{l}$ should be mapped to. 
% More specifically, the block $({\bf{W}}_{l})_{ij}$ has the following structure:

% \begin{equation}
% \label{circulant_matrix}
% \mathcolorbox{yellow}{\mathbf({\bf{W}}_{l})_{ij}
% =
% \begin{bmatrix}
%  c_{1}      & c_{L_b}      & \cdots & c_{{3}}  & c_{2}      \\
%  c_{2}      & c_{1}      & \cdots  & c_{{4}} & c_{{3}}      \\
%  \vdots & \vdots & \ddots & \vdots  & \vdots\\
%  c_{L_{b-1}}      & c_{L_{b-2}}     & \cdots & c_{1} & c_{L_b}      \\
%      c_{L_b}      &  c_{L_{b-1}}     & \cdots & c_{2}  & c_{1}      \\
% \end{bmatrix},}
% \end{equation}
% \hl{}
We introduce auxiliary variables ${\bf{Z}}_{l}$ and ${\bf{U}}_{l}$, which have the same dimensionality as ${\bf{W}}_{l}$. Through the application of ADMM\footnote{{The details of the ADMM algorithm are discussed in \cite{boyd2011distributed,zhang2018systematic}. We omit the details because of space limitation.}}, the original structured training problem can be decomposed into two subproblems, which are iteratively solved until convergence. In each iteration $k$, the first subproblem is
\vspace{-0.6cm}
\begin{equation}
\label{4}
 {\underset{ \{{\bf{W}}_{l}\}}{\text{minimize}}
\ \ \ f \big( \{{\bf{W}}_{l} \}_{l=1}^N\big)+\sum_{l=1}^{N} \frac{\rho_{l}}{2}  \| {\bf{W}}_{l}-{\bf{Z}}_{l}^{k}+{\bf{U}}_{l}^{k} \|_{F}^{2},} \\
\end{equation}
where ${\bf{U}}_{l}^{k}$ is the dual variable updated in each iteration, ${\bf{U}}_{l}^{k}:={\bf{U}}_{l}^{k-1}+{\bf{W}}_{l}^{k}-{\bf{Z}}_{l}^{k}$. In the objective function of (\ref{4}), the first term is the differentiable loss function of RNN, and the second quadratic term is differentiable and convex. As a result, this subproblem can be solved by stochastic gradient descent and the complexity is the same as training the original RNN. A large number of contraints are avoided here. The result of the first subproblem is denoted by ${\bf{W}}_{l}^{k+1}$.Proven in \cite{boyd2011}, the global optimal solution of the second subproblem is to find a Euclidean mapping of ${\bf{W}}_{l}^{k+1}+{\bf{U}}_{l}^{k}$ to the closest structured (circulant) matrix format. The result of the second subproblem is denoted by ${\bf{Z}}_{l}^{k+1}$.

\begin{figure}[t]
\begin{center}
\includegraphics[width = 0.3\textwidth]{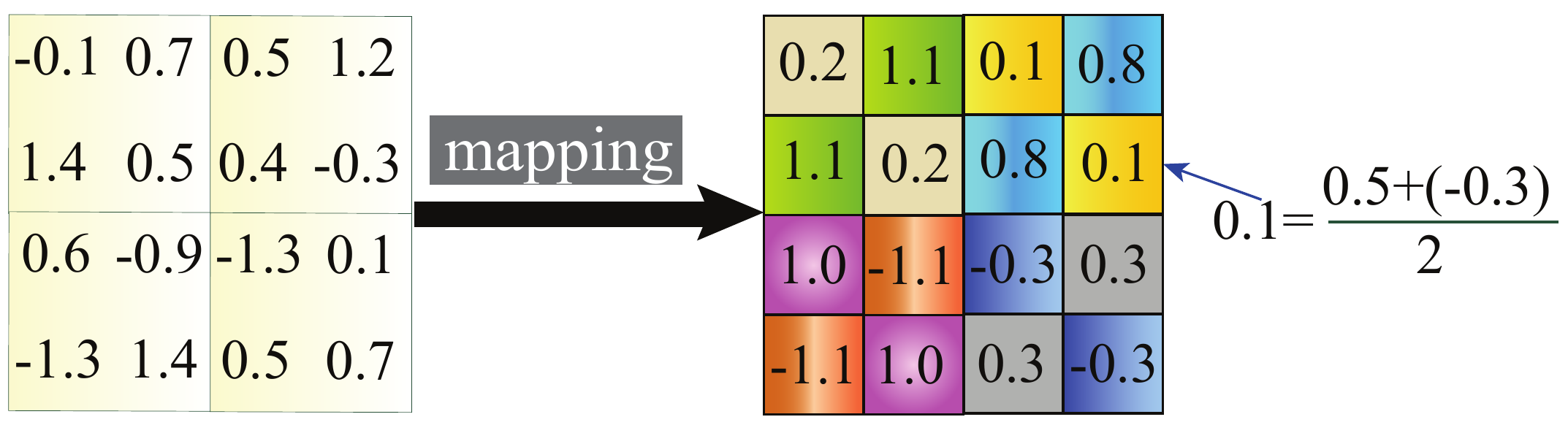}
% \vspace{-0.6cm}
\caption{{Euclidean mapping for a 4 $\times$ 4 matrix with block size of 2.}}
\label{fig:admm_exmp}
\end{center}
% \vspace{-0.6cm}
\end{figure}

For better illustration, let $({\bf{W}}_{l}^{k+1}+{\bf{U}}_{l}^{k})$ denote a specific matrix to be mapped, and let $({\bf{Z}}_{l}^{k+1})$ denote the corresponding structured format. For the $ij^{th}$ block,
% as shown in the structured format Eqn. (\ref{circulant_matrix}),
the elements $(1,1)$, $(2,2)$,..., $(L_b,L_b)$ of $({\bf{Z}}_{l}^{k+1})_{ij}$ should be equal. For Euclidean mapping, we have:
% \begin{align}
\begin{equation} \label{average}
\begin{aligned}
% \begin{split}
   & {({\bf{Z}}_{l}^{k+1})_{ij,(1,1)}=({\bf{Z}}_{l}^{k+1})_{ij,(2,2)}=...=({\bf{Z}}_{l}^{k+1})_{ij,(L_b,L_b)}} \\
     &   {=\frac{({\bf{W}}_{l}^{k+1}+{\bf{U}}_{l}^{k})_{ij,(1,1)}+...+({\bf{W}}_{l}^{k+1}+{\bf{U}}_{l}^{k})_{ij,(L_b,L_b)}}{L_b}}
\end{aligned}
% \end{split}
\end{equation}
{Similarly the other entries in $({\bf{Z}}_{l}^{k+1})_{ij}$ can be calculated. We have proved that this is the optimal analytical solution of the second subproblem. Fig.~\ref{fig:admm_exmp} illustrates an example of the Euclidean mapping by applying Eqn.~(\ref{average}).}

\begin{figure} [t]
 	\centering
%  	\vspace{-1.2em}
 	\includegraphics[width=0.85\columnwidth]{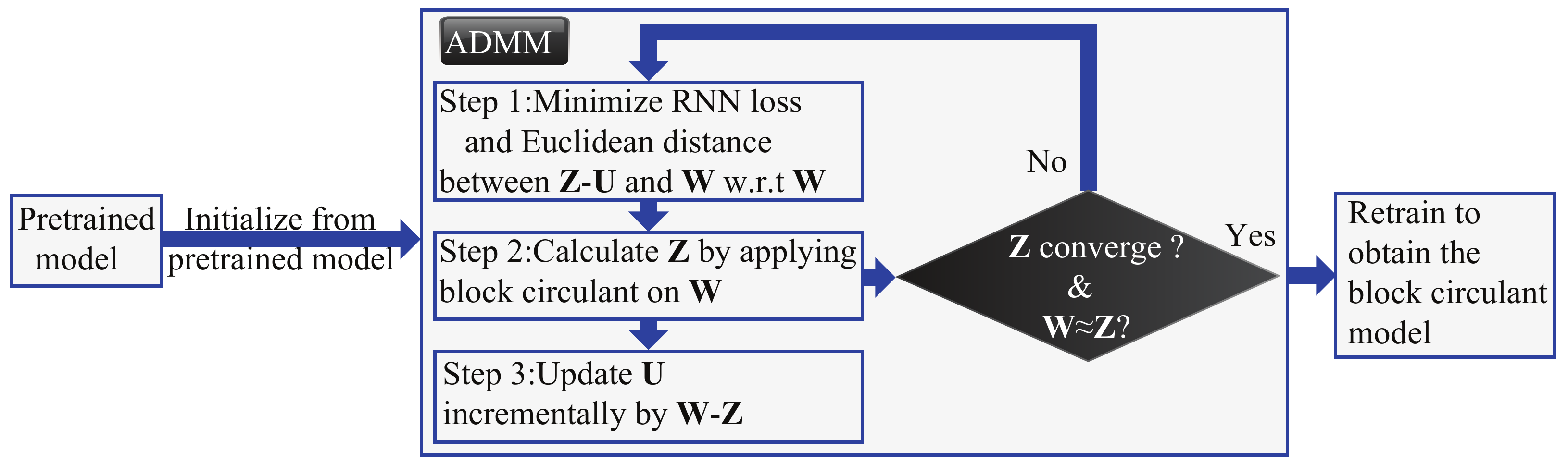}	
 	\caption{{The overall procedure of ADMM-based structured matrix training.}}
 	\label{fig:ADMM}
 \end{figure}

{The overall procedure of ADMM-based structured matrix training is shown in Fig.~\ref{fig:ADMM}. Essentially speaking, it iteratively (i) map ${\bf{Z}}_{l}^{k+1}$ to the structured format in the optimal manner, and (ii) use the mapped ${\bf{Z}}_{l}^{k+1}$ as a dynamic regularization target for weight training. 
Upon convergence the RNN weights will converge to the structured format. 
The proposed method effectively overcomes the limitation of combinatorial constraints and achieves higher training accuracy compared with the prior work, as shall be seen in experimental results. }

\section{RNN Model Design Exploration: A Top-Down View }
\label{sec:model1}

% In this section, we explore two ASR applications using RNN model: TIMIT and Librispeech.
In this section, we perform RNN model design exploration at the algorithm level, in order to shed some light on RNN training trial reductions. More specifically, we provide an analysis of the effect of model type (LSTM or GRU), layer size, and block size on the overall accuracy. The design variable with the least impact on the overall accuracy should be given priority in design optimization. We focus on TIMIT benchmark, the most widely utilized benchmark for ASR applications. In the following, we will provide a detailed discussion on the data set, RNN models, and results and observations. 
% We have shared the anonymous link of the related codes and information online\footnote{\href{https://github.com/AnonymousAccount1313/BlockCIrculantRNN}{{https://github.com/AnonymousAccount1313/BlockCIrcula\\ntRNN}.}}.

\textbf{Dataset.} 
The TIMIT dataset \cite{garofolo1993darpa} contains broadband recordings of $630$ speakers of eight major dialects of American English, each reading ten phonetically rich sentences, totally $6,300$ utterances. The TIMIT corpus includes time-aligned orthographic, phonetic and word transcriptions as well as a 16-bit, 16kHz speech waveform file for each utterance. 

\textbf{RNN Models.}
The RNN models utilized in the design exploration are summarized in Table~\ref{tbl:LSTM_accuracy} and Table~\ref{tbl:GRU_accuracy}. We stack multiple RNN layers to build our network.
The number of layers and layer sizes (dimensionality of $\mathbf{c}_t$) are listed in the tables.
For an LSTM cell, $256-256-256$ means that the network has three layers of LSTM cells with $256$ hidden neurons in $\mathbf{c}_t$.
The block sizes (as a power of 2) are listed in the same format as layer sizes correspondingly.
``$-$'' means that we do not apply (block-)circulant matrix on the network, which is the baseline model for that specific network structure. The baseline model with layer size 1,024 is {the same as the baseline in ESE}~\cite{han2017ese}. 
We also list the configuration options like ``peephole'' and ``projection''.
The performance is evaluated by \emph{phone error rate} (PER) or \emph{word error rate} (WER) and degradations compared to the corresponding baseline model. The smaller the PER or WER, the better of the corresponding RNN model.

\begin{table}[t]
	\centering
	\vspace{-0.1em}
	\caption{Comparison among LSTM based RNN models}
	\vskip -0.8em
	\label{tbl:LSTM_accuracy}
	\resizebox{1.0\columnwidth}{!}{
		\begin{tabular}{c|c|c|c|c|c|c}
			\hline
			\multirow{2}{*}{ID}  & Layer &Block &Peep- & Projection & Phone Error & PER degra-\\
		 	& Size  & Size        &  hole                   &         ($512$)                    & Rate (PER) \% & dation (\%)\\\hline
			1 &$256-256-256$       & $-$ &      $\times$   &     $\times$ &20.83&  $-$ \\\hline
			2 &$256-256-256$       & $2-2-2$ &  $\times$   &     $\times$ &20.75& $-0.08$  \\\hline
			3 &$256-256-256$       & $4-4-4$ &  $\times$   &     $\times$ &20.85& $0.02$  \\\hline\hline
			4 &$512-512    $       & $-$   &    $\surd$    &     $\times$ &20.53&  $-$ \\\hline
			5 &$512-512    $       & $4-4$ &    $\surd$    &     $\times$ &20.57& $0.04$  \\\hline
			6 &$512-512    $       & $4-8$ &    $\surd$    &     $\times$ &20.85& $0.28$  \\\hline
			7 &$512-512    $       & $8-4$ &    $\surd$    &     $\times$ &20.98& $0.41$  \\\hline
			8 &$512-512    $       & $8-8$ &    $\surd$    &     $\times$ &21.01& $0.48$  \\\hline\hline
			9 &$1024-1024 $       & $-$   &    $\surd$    &     $\surd$  &20.01&  $-$  \\\hline
			10&$1024-1024 $       & $4-4$ &    $\surd$    &     $\surd$  &20.01&  $0.00$ \\\hline
			11&$1024-1024 $       & $4-8$ &    $\surd$    &     $\surd$  &20.05&  $0.04$ \\\hline
			12&$1024-1024 $       & $8-4$ &    $\surd$    &     $\surd$  &20.10&  $0.09$ \\\hline
			13&$1024-1024 $       & $8-8$ &    $\surd$    &     $\surd$  &20.14&  $0.13$ \\\hline
			14&$1024-1024 $       & $8-16$&    $\surd$    &     $\surd$  &20.22&  $0.21$ \\\hline
			15&$1024-1024 $       & $16-8$&    $\surd$    &     $\surd$  &20.29&  $0.28$ \\\hline
			16&$1024-1024 $       & $16-16$ &  $\surd$    &     $\surd$  &20.32&  $0.31$ \\\hline
		\end{tabular}
	}
\end{table}

\begin{table}[t]
	\centering
	\caption{Comparison among GRU based RNN models}
	\vskip -0.8em
	\label{tbl:GRU_accuracy}
	\resizebox{0.85\columnwidth}{!}{
		\begin{tabular}{c|c|c|c|c}
			\hline
			\multirow{2}{*}{ID}  & Layer &Block  & Phone Error & PER\\
		 	& Size  & Size                         & Rate (PER) \% & degradation (\%)\\\hline
			1 &$256-256-256$       & $-$  &20.72&  $-$ \\\hline
			2 &$256-256-256$       & $4-4-4$  &20.81& 0.09  \\\hline
			3 &$256-256-256$       & $8-8-8$  &20.88& 0.16  \\\hline\hline
			4 &$512-512    $       & $-$    &20.51&  $-$ \\\hline
			5 &$512-512    $       & $4-4$  &20.55& 0.04  \\\hline
			6 &$512-512    $       & $4-8$  &20.73& 0.22  \\\hline
			7 &$512-512    $       & $8-4$  &20.89& 0.38  \\\hline
			8 &$512-512    $       & $8-8$  &20.95& 0.44  \\\hline\hline
% 			9 &$512-512    $       & $16-16$  &21.34& 0.83  \\\hline\hline
			9 &$1024-1024 $       & $-$     &20.02&  $-$ \\\hline
			10 &$1024-1024 $       & $4-4$   &20.03& 0.01  \\\hline
			11 &$1024-1024 $       & $4-8$   &20.08& 0.06   \\\hline
			12 &$1024-1024 $       & $8-4$   &20.13& 0.11  \\\hline
			13 &$1024-1024 $       & $8-8$   &20.20& 0.18  \\\hline
			14 &$1024-1024 $       & $8-16$  &20.25& 0.23  \\\hline
			15 &$1024-1024 $       & $16-8$  &20.31& 0.29  \\\hline
			16 &$1024-1024 $       & $16-16$  &20.36& 0.33  \\\hline
		\end{tabular}
	}
\end{table}

% \subsubsection{TIMIT}

\textbf{Results Discussion and Observations.} From Table~\ref{tbl:LSTM_accuracy} and Table~\ref{tbl:GRU_accuracy}, we can observe that the block-circulant matrix-based framework results in very small accuracy degradation compared with the baseline model. More specifically, when the block size is 4 (4 $\times$ parameter reduction) or smaller, there is in general no accuracy degradation compared with the corresponding baseline. When the block size is 8 (8 $\times$ parameter reduction), the accuracy degradation is negligible, around 0.1\%-0.15\%. When the block size is 16, the accuracy degradation is still only around 0.3\%. As discussed before, the baseline model with layer size 1,024 is the same as the baseline in ESE~\cite{han2017ese}. Then we can conclude that the block-circulant matrix-based framework outperforms ESE in terms of model compression. This is because ESE achieves 9$\times$ parameter reduction with 0.3\% accuracy degradation. This parameter reduction even does not account for the indices, which are needed at least one for each parameter in the network structure after pruning. We will observe in the hardware experimental results that the performance and energy efficiency gains are even more significant compared with ESE, thanks to the regularity in this framework.

Moreover, the above design exploration procedure provides observations on the RNN model selection and optimization, which could shed some lights on training trial reductions. We can observe that changing from LSTM to GRU or using a block size of 4 or smaller will not result in accuracy degradation. Therefore, if the accuracy requirement is very tight for the target application, we can in general change to GRU and/or using a block size of 4. In this way the amounts of computation and storage are reduced, which is directly related to the performance and energy consumption in hardware implementations, with zero accuracy degradation. If a small amount of accuracy degradation is allowed, then the top priority is using a block size of 8 or 16 compared with a smaller LSTM/GRU RNN model (i.e., a smaller layer size). This is because that the block-circulant matrix based framework, as shown in the two tables, results in smaller amount of accuracy loss and greater computation/storage reduction compared with a smaller LSTM/GRU RNN model. For ASR applications, a block size of 8 or 16 will make the whole RNN model easily accommodated by the on-chip BRAM of FPGAs. This observation validates the effectiveness of the block-circulant framework, and becomes the basis for reducing RNN training trials in the overall design optimization procedure to be discussed in Section \ref{sec:phase1}.

\subsection{The Underlying Principle of Observation}

A natural question to ask is: what is the underlying reason that using a larger block size (or more generally, reducing weights) results in smaller accuracy degradation compared with reducing the layer size? The reason is that the number of weights exhibits a higher degree of redundancy compared with the number of hidden neurons (the former is in the order of O($n^2$) whereas the latter is in the order of O($n$)). Therefore, reducing the number of weights typically results in very minor accuracy degradation, or no degradation at all, compared with reducing layer size. This observation is also discovered in~\cite{han2015deep,han2016eie}. Besides, the overfitting issue can be partially mitigated and the generality of RNN can be improved through weight reductions.

\section{RNN Model Design Exploration: A Bottom-Up View }
\label{sec:model2}
In this section, we perform the second RNN model design exploration focusing on computation reductions. More specifically, we analyze {the amount of computation in each layer as a function of block size, accounting for various techniques for computation reductions.} It can effectively set a proper range of block size optimization, thereby facilitating the overall design optimization.

\subsection{Techniques for Computation Reduction in the Block-Circulant Framework}

\subsubsection{FFT-IFFT Decoupling}
\label{sec:decoup}
% In the block-circulant framework, 
We can pre-calculate FFT$\mathbf{(w}_{ij})$  vectors and store them in BRAM before the inference phase since all the weights are fixed after the training process. From Eqn.~(\ref{eqn:FFT}), we observe that the calculations of FFT$\mathbf{(x}_{j})$ and IFFT are always executed in pairs.
% Therefore, we can combine and implement them as FFT$\mathbf{}/\mathbf{IFFT}$ kernel. 
There are $N$ multipliers between FFT and IFFT, which calculate the dot product of the intermediate results of FFT$\mathbf{(x}_{j})$ and weight values FFT$\mathbf{(w}_{ij})$ pre-stored in BRAM.  

% Each row of weight matrix (pre-calculated $\mathbf{w}_{ij}$) needs to element-wise multiply by the input feature minibatch fetched from BRAM.

% We observe that the intermediate results FFT$\mathbf{(x)}_{j}$ need to be utilized to calculate all
% the $\mathbf{a}_i$. We propose that instead of constantly operating the FFT calculation for each $\mathbf{a}_i$, we could pre-calculate FFT$\mathbf{}(\mathbf{x}_j)$ vectors, and store them in BRAM. Thus, for each $\mathbf{a}_i$, we can effectively re-use the pre-calculated FFT$\mathbf{}(\mathbf{x}_j)$ vectors. By
% performing such pre-calculation of FFT$\mathbf{}(\mathbf{x}_j)$, the total
% amount of FFT operations needed to calculate $\mathbf{a=Wx}$ can be reduced from
% $p\cdot q$ to $q$.

\begin{figure} [b]
 	\centering
%  	\vspace{+0.5em}
 	\includegraphics[width=0.9\columnwidth]{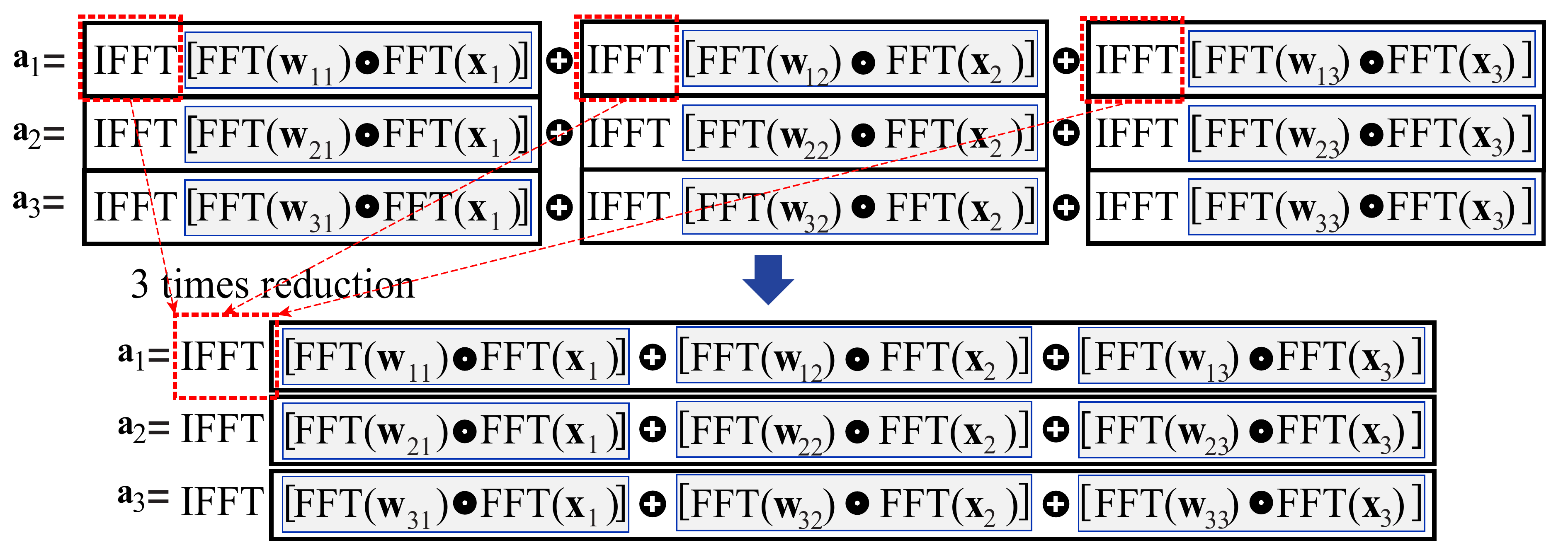}
 	\caption{A demonstration of matrix-vector multiplication (matrix size 3 $\times$ 3 blocks) (top); and the calculation process using decoupling techniques (bottom). FFT$\mathbf{}(\mathbf{w}_{ij})$'s are pre-calculated and stored in BRAM.}
%  	{This figure should be made clearer, maybe change to 3 by 3, and the bold font of FFT/IFFT throughout this paper should better be changed.}}
 	\label{fig:Decoup}
 \end{figure}

To further achieve a higher degree of parallelism, we adopt the \emph{FFT/IFFT decoupling} technique concentrating on reducing the number of FFT/IFFT computations. 
% Suppose a weight matrix has $p \times q$ blocks after partitioning with block size $N$. $N$ is also the input size of FFT/IFFT kernel.
We give a demonstration with weight matrix size $3 \times 3$ blocks shown in Fig.~\ref{fig:Decoup}, in which each input has 3 blocks (segments). The intermediate results FFT$\mathbf{(x}_{1})$ need to be utilized 3 times to finish the calculation process for 3 output segments. We propose to pre-calculate FFT$\mathbf{(x}_{1})$, and store the intermediate results in BRAM. Thus, for each $\mathbf{a}_i$, we can effectively re-use the pre-calculated FFT$\mathbf{(x}_{1})$ vector. Additionally, according to~\cite{oppenheim1999discrete}, FFT/IFFT are linear functions. Thus, FFT/IFFT can be decoupled and IFFT will be executed after the accumulation. For a weight matrix with $p \times q$ blocks, the FFT$(\mathbf{x}_j)$ pre-calculation could reduce the number of FFT calculations from $p\cdot q$ to $q$, and the FFT/IFFT decoupling could also reduce the number of IFFT from $p\cdot q$ to $p$.

\subsubsection{Leveraging Special Property of Real-Valued FFTs}
\label{sec:real valued}

We perform further computation reduction making use of the following observation: 
The inputs/outputs of each layer are \emph{real values} without imaginary parts in actual RNN applications. We focus especially on multiplications since they are more expensive to implement than additions in hardware. For example, 
both $\mathbf{x}_j$ and $\mathbf{w}_{ij}$ are real-valued vectors. Computation reductions are achieved in three aspects. First, 
FFT/IFFT can be simplified because the result of FFT with real-value inputs will be symmetric in real and imaginary parts except for the base component \cite{salehi2013pipelined,chang2003efficient}. As a result the last level of butterfly plot~\cite{cooley1965algorithm} in FFT computation and the first level of IFFT can be reduced by half. 
Second, the multiplication computation of FFT$\mathbf{(x}_{j})\odot$FFT$\mathbf{(w}_{ij})$ (and corresponding accumulations), along with storage of intermediate results, are also reduced by half. This is also the result of the symmetric property. The second aspect is even more important because element-wise multiplications/additions 
will become the dominant computing part.

Finally, further computation reduction is achieved in FFT/IFFT leveraging the FFT/IFFT properties. Take the FFT as an example, the first two levels in the butterfly plot of FFT do not need to perform multiplication because the $W$ twiddle factors are 1, -1, $i$, or $-i$ in these two levels. Only half of butterfly units in the third level need to perform multiplication calculation; only $1/4$ in the fourth level, $1/8$ in the fifth level, and so on. Reducing the number of multiplications will be critical to the overall design optimization.

\subsection{Observation and Discussions}
 \begin{figure} [t]
 	\centering
 	\vspace{-1.2em}
 	\includegraphics[width=0.9\columnwidth]{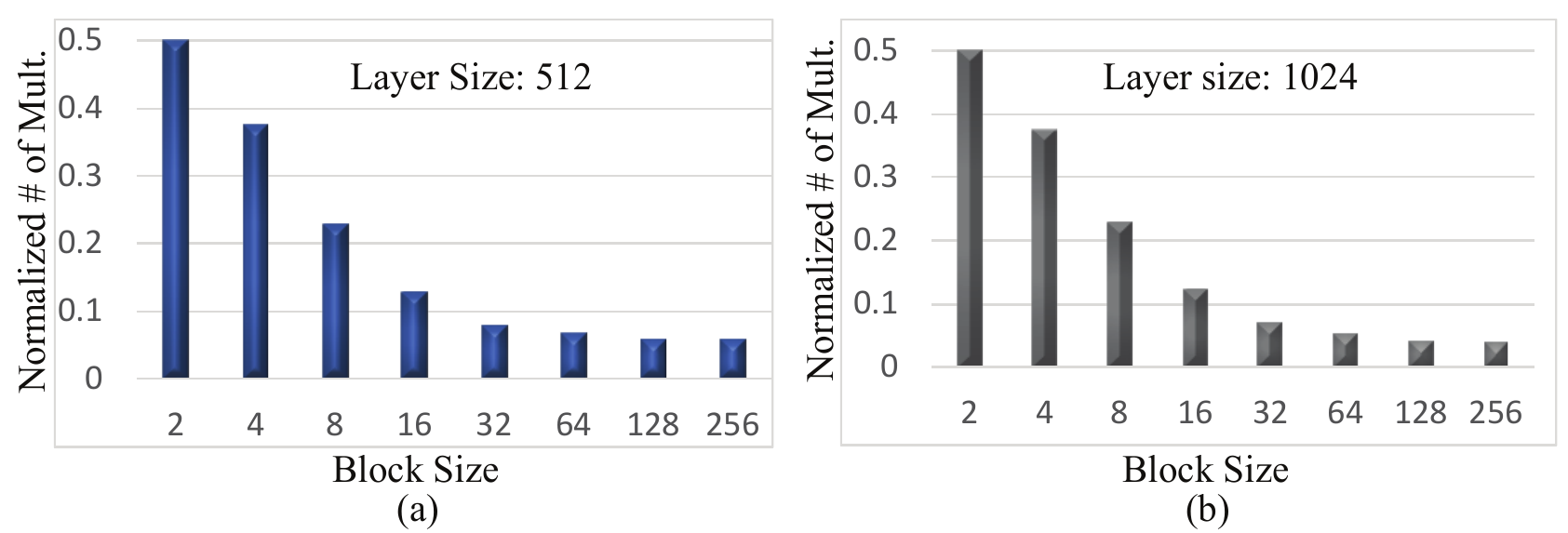}	
 	\vskip -0.5em
 	\caption{ Normalized number of multiplications as a function of block size with (a) layer size 512 and (b) layer size 2014. }
 	\label{fig:layersize}
 \end{figure}
Accounting for the above-mentioned computation reduction techniques, we analyze the amount of computation in an RNN layer as a function of block size. We consider layer sizes 512 and 1024 that are typical for ASR applications. Fig.~\ref{fig:layersize} illustrates the amount of multiplication computation (which is more expensive in hardware than additions) as a function of block size with these two layer sizes. The multiplications are normalized by the initial amount with block size 1 (i.e., without application of block-circulant matrices). Please note that the block size is a power of 2 as mentioned above.

As can be observed, the computation reduction will converge when the block size reaches 32 or 64, and the amount of computation can even increase when we further increase the block size. The reason is because the increase in computation in FFT/IFFT will compensate with the decrease in element-wise multiplications.
As the accuracy will also degrade when block size reaches 32 or 64, we can set a upper bound of 64 (or 32) of block size, thereby facilitating the overall design optimization.

\section{E-RNN Framework: Phase I}
\label{sec:phase1}

\subsection{Overview of the E-RNN Framework}

Based on the above two design explorations and corresponding observations, we present the E-RNN design optimization framework of RNN implementations in FPGA. The optimization objectives are performance and energy efficiency under the overall accuracy requirement. 
The optimization variables include model type (LSTM, GRU, etc.), block size and layer size optimization, hardware implementation structure and parallelism degree, quantization and activation functions, etc. 

To facilitate the design optimization procedure, we divide the overall design optimization into two phases. \emph{Phase I} lies at the interface between algorithm and hardware and determines RNN model specifications, including model type, layer size, and block size, under the overall accuracy constraint. The objective is to reduce the RNN model size and computations. \emph{Phase II} focuses on hardware-oriented optimization given the RNN model, and determines the hardware structure, the number of \emph{processing elements} (PEs), quantization scheme and activation function implementations, etc.

\subsection{E-RNN Phase I: Deriving the RNN Model}

% {Yanzhi: A figure of E-RNN Phase I should go here.}

%  \begin{figure} [t]
%  	\centering
%  	\vspace{-1.2em}
%  	\includegraphics[width=0.7\columnwidth]{figs/fig_phase1.pdf}	
%  	\vskip -0.5em
%  	\caption{ The Phase-I algorithm of E-RNN. }
%  	\label{fig:phase1}
%  \end{figure}

The Phase-I algorithm of E-RNN framework is illustrated in Fig.~\ref{fig:phase1}. It consists of three major steps, \emph{initial sanity check}, \emph{block size optimization}, and \emph{fine tuning}. Clearly this algorithm has made use of the first observation that block size optimization should be prioritized over layer size. The second observation on block size range is effectively utilized in the second step to reduce RNN training trials. The objective of Phase I is to reduce the RNN model size storage and computations (please note that computation will be the primary goal of optimization as long as the whole RNN model fits into BRAM of FPGA), and the overall accuracy constraint needs to be satisfied.

The \emph{Step One} performs a sanity check on whether it is possible to accommodate the whole RNN model using on-chip BRAM. As the block size should be the primary optimization variable, we start from the LSTM RNN baseline model due to its high reliability, and estimate the block size required to fit into BRAM. 
For example, the FPGAs we test on (Xilinx Kintex UltraScale or Virtex-7) have 4-8MB BRAM. For the ASR application and LSTM/GRU model, a block size of 4 or 8 will fit the whole RNN model into BRAM. A block size 8 will be safer in order to allocate certain portion of BRAM for inputs/outputs. 
% \hl{On the other hand, when a LSTM/GRU  model is too large to fit the on-chip BRAM for certain platforms such as Zynq-7000 with BRAM size of 256KB, the off-chip memory (DDR) access is necessary. The Zynq-7000 development board can afford at least 1GB DDR3 memory, which is much bigger than the BRAM, and is enough to load the LSTM/GRU  model. Please note that the off-chip memory access will not change the hardware design and only cause memory access overhead and overall performance degradation. Therefore, we prefer to use on-chip memory since the data transfers between the processor and off-chip memory consume more power, at a relatively slow speed. 
% % The ability to accommodate the whole RNN model on on-chip BRAM is one of the reasons R-ENN can outperform ESE more than 20X in both performance and energy efficiency at the similar accuracy loss level.
% }
The required block size serves as a lower bound for the subsequent step. 

As long as the whole RNN model fits into the on-chip BRAM of FPGA, the primary goal of optimization in Phase I should be computation reduction rather than storage, because the former is directly correlated with performance/energy efficiency of hardware implementation. As a result, computation reduction becomes the primary goal of \emph{Step Two} (block size optimization). Remind that we have derived the lower bound of block size from Step One and the upper bound from Section~\ref{sec:model2}. In Step Two, we find the largest block size within these bounds that satisfy the overall accuracy constraint. With both bounds and the fact that the block size should be a power of 2, the number of RNN training trials can be significantly reduced.
For example, if the lower bound is 8 and the upper bound is 32 (or 64), there are at most 3 or 4 training trials needed for block size optimization.

Up till now we are using LSTM RNN model and have derived a desirable block size. In \emph{Step Three} (fine tuning), we determine the model type (LSTM or GRU) and perform fine tuning of block size (allowing for a larger block size for relatively unimportant weight matrices). Determining the model type is straightforward. We simply change from LSTM to GRU with block size fixed (the GRU model will be fitted into BRAM because it is smaller than LSTM), and perform a single RNN training. If the accuracy requirement can still be satisfied, it is desirable to shift from LSTM to GRU because of less computation and storage. In the ASR applications, we can switch safely from LSTM to GRU without accuracy loss.

In this step, we will also increase the block size for relatively unimportant weight matrices, which will not cause a significant accuracy degradation. Those weight matrices include the input and output matrices that will not propagate from each time $t$ to the subsequent time step. As indicated in~\cite{ding2017circnn}, supporting multiple block sizes is achieved thanks to the recursive property of FFTs~\cite{salehi2013pipelined,chang2003efficient} with proper control mechanism. In order to limit the number of additional RNN training trials and simplify the control mechanism, we limit the maximum type of block sizes to be 2. In other words, we will only use a single larger block size for the input and
output matrices. The number of additional trainings will be 1 or 2 accounting for the upper limit of block size from Section~\ref{sec:model2}. 
In our actual experiments, if the block size is 8 (or 16), there is only need for a single test of block size 16 (or 32) for input/output matrices, since a larger block size will result in accuracy degradation.

In summary, the total number of training trials is limited to around 5 thanks to the two observations in Section~\ref{sec:model1} and Section~\ref{sec:model2}. This number becomes affordable for ASR and many other applications.

\section{E-RNN Framework: Phase II}

Given the RNN model generated by Phase I, \emph{Phase II} focuses on hardware-oriented optimization, and determines the hardware implementation structure, \emph{processing elements} (PEs) design, quantization scheme and activation function implementations, etc.

\subsection{E-RNN Hardware Architecture}

\begin{figure}[t]
    \centering
    \includegraphics[width=0.65\columnwidth]{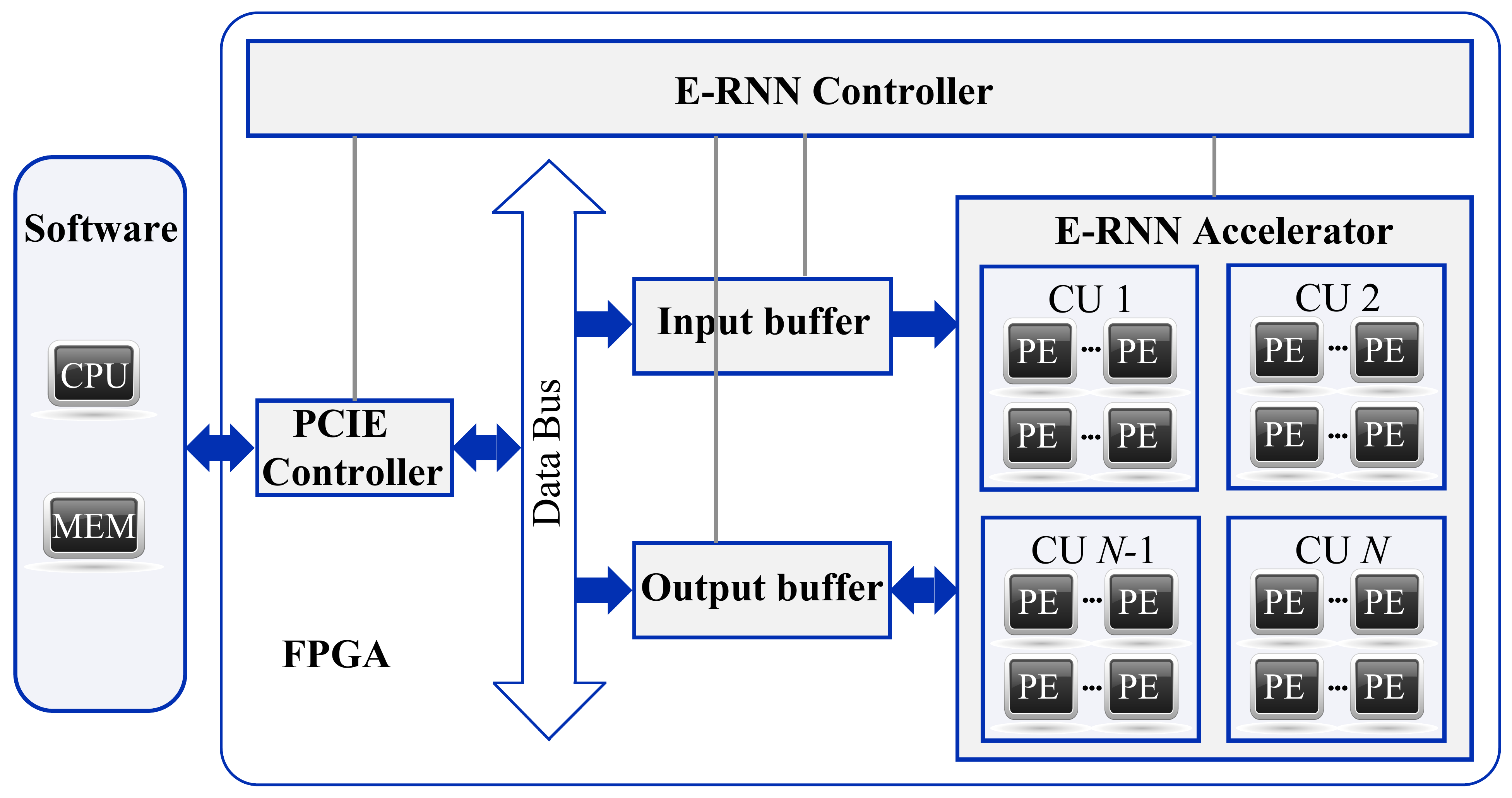} 
    \vskip -0.5em
    \caption{The overall E-RNN hardware architecture.}
    \label{fig:sys_overall}
 \end{figure}

Fig.~\ref{fig:sys_overall} demonstrates the E-RNN hardware architecture. 
% The whole system consists of two parts: the software program on CPU and the E-RNN hardware accelerator on an FPGA chip.
A CPU and a host memory communicate with the FPGA chip through PCI-Express (PCIE) bus. They can transmit the input voice vector to the FPGA and receive the computation results from the accelerator on FPGA. The host memory initially stores all the parameters (weight matrices and biases) and input voice vectors, which will be further loaded into on-chip memories (BRAM) of FPGA for online inference.

% The proposed E-RNN method offers sufficient memory capacity to store all the parameters and voice vectors on the on-chip BRAM of the selected FPGA development board. Therefore, we can save the extra accessing latency on external memory (DRAM) of FPGA since the latency of DRAM is much higher than the latency of BRAM.  

In the FPGA chip, we implement the E-RNN controller, E-RNN accelerator, PCIE controller, and input/output buffer. The E-RNN accelerator comprises a group of \emph{processing elements} (PEs). PEs are the basic computation block for one set of input voice vectors with the corresponding weights and are primarily responsible for the computing tasks in LSTM and GRU. A handful of PEs and their peripheral components are bundled as a \emph{compute unit} (CU). Each CU implements the LSTM/GRU model and computes one input voice vector sequence independently. The E-RNN controller takes charge of the process of data fetching of the PCIE controller. Most importantly, it determines the computation pipeline flow of the whole LSTM/GRU network. The on-chip input buffer and output buffer have the data ready for PEs and collect the output results from the accelerator. The E-RNN accelerator fetches parameters and input voice vectors from on-chip BRAM and collects the results and writes back to BRAM.

\subsection{PE Design}
  \begin{figure}[t]
    \centering
    \includegraphics[width=0.6\columnwidth]{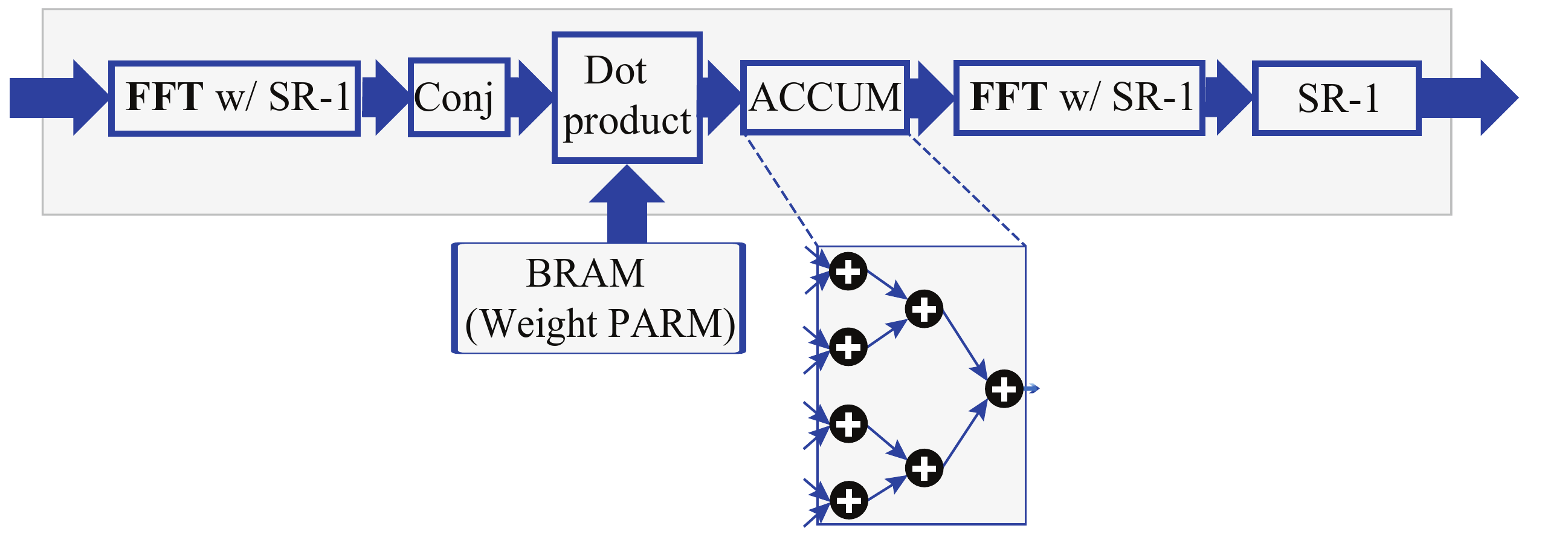} 
    \vskip -0.5em
    \caption{The PE design in FPGA implementation.}
    \label{fig:PE}
 \end{figure}
 
As shown in Fig.~\ref{fig:PE}, a PE consists of two FFT operators, $M$ multipliers, a conjugation operator, $\log_2 N$ right shifting registers, and an accumulator. The accumulator is an adder tree with $N$ inputs (same as the FFT size). 
% Given the total number of $T$ digital signal processing (DSP) slices of an FPGA chip, the number of PEs can be calculated as: $\#PE=T/M$. 
Due to the resource limitation on FPGAs, we need to let PEs operate using time-division multiplexing (TDM) for different blocks.
{
% The resource usage model including Look-up tables (LUTs), DSP blocks, and BRAM of an FPGA implementation can be mathematically described. 
Suppose the DSP and LUT usage of one PE are $\Delta DSP$ and $\Delta LUT$, respectively. 
% The models of \# DSP, \# LUT, and \# BRAM are shown as follows,
The number of PEs can be expressed as: $\# PE= min\{\floor {\frac{DSP}{\Delta DSP}}, \floor {\frac{LUT}{\Delta LUT}}\}$, where $DSP$, $LUT$ are the total resources of  DSP and LUT, respectively.
}

% \begin{equation}
% \label{eqn:pe}
% \$mathcolorbox{yellow}{\# PE= min\{\floor {\frac{DSP}{\Delta DSP}}, \floor {\frac{LUT}{\Delta LUT}}\}}
% \end{equation}

% \hl{where $\#CONV_{D}$, $\#CONV_{L}$ are the number of CONV operations for DSP and LUT, respectively. Generally, in Xilinx Virtex-7 FPGA fabric, the BRAM size is 36kb and the BRAM bandwidth is 64b.}

%, where $M=N+\log_2 N$
% To increase the number of PEs, we need to reduce the number of multipliers $M$ (.i.e., PE size).

% We adopt an end-to-end optimization framework including hardware system design, PE optimization, and quantization (from 16 bit in C-LSTM to 12 bit in E-RNN). The PE level optimization reduces a large number of multipliers and adders and achieves a higher degree of computation parallelism, which leads to the significant improvement of performance and energy efficiency in the FPGA implementation. Among the three, the first two components are more effective compared to quantization: reducing from 16 bit to 12 bit only accounts for less than 10% performance improvement. 

\subsection{Compute Unit (CU) Implementation}

% The CU implementations of LSTM and GRU are determined by the data dependencies and flows. The complexity of matrix-vector multiplication is much higher than that of point-wise multiplications and additions. Therefore, we need to separate these operations based on the dependencies, and fuse other low complexity operations carefully. 

  \begin{figure}[b]
    \centering
    \includegraphics[width=0.65\columnwidth]{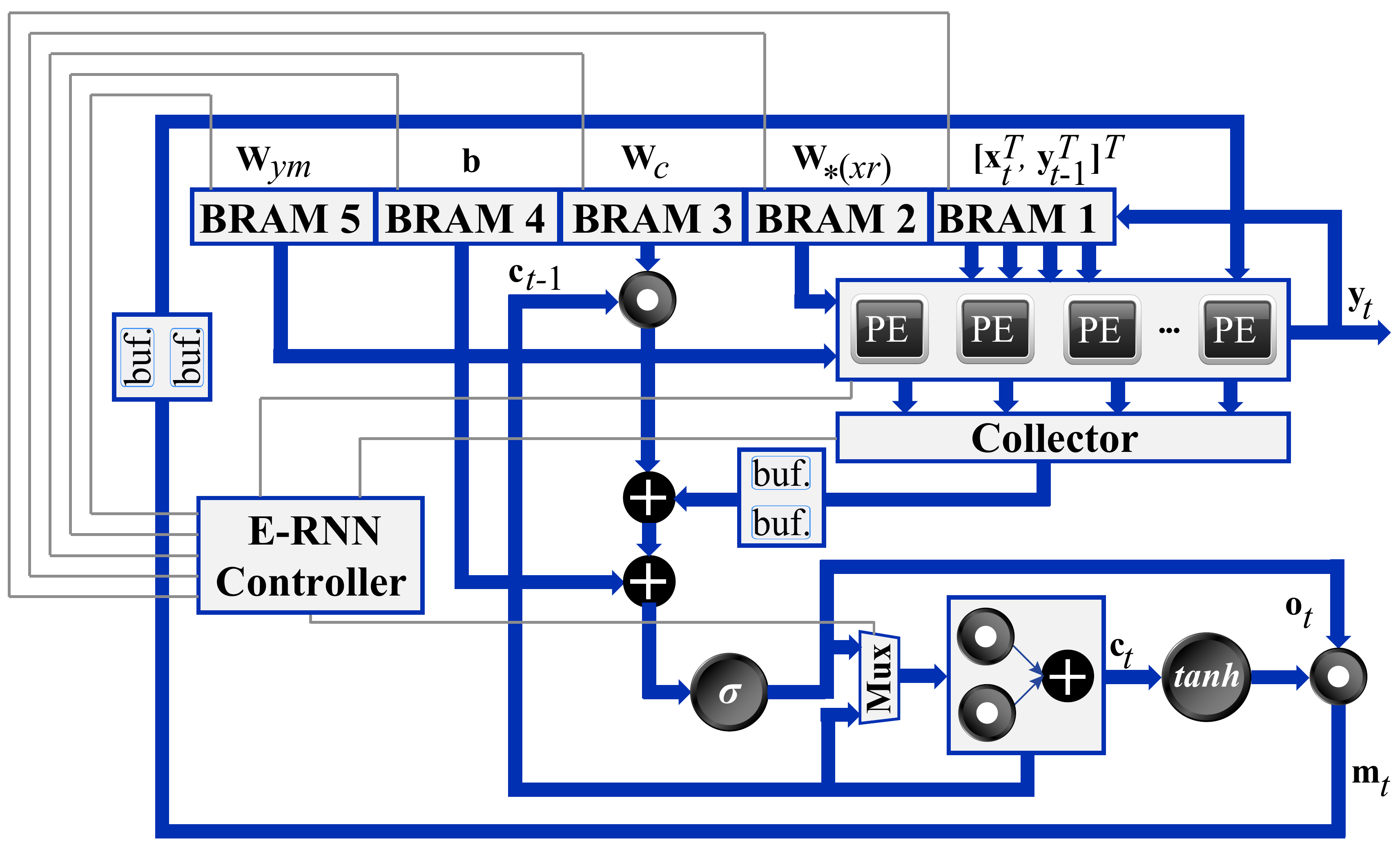} 
    \vskip -0.5em
    \caption{One compute unit (CU) with multiple processing elements (PEs) of LSTM.}
    \label{fig:lstm_arch}
 \end{figure}

\subsubsection{CU implementation of LSTM}
 
The proposed CU architecture for LSTM model described in Eqn. (\ref{eqn:model}) can be implemented using above designs, shown in Fig.~\ref{fig:lstm_arch}. The architecture consists of multiple PEs, sigmoid/tanh, double buffers, and multiplier-adder block.
% We adopt fine-grained pipelining technique to implement the matrix-vector multiplication. 
There are five BRAM blocks. BRAM 1 stores input features. The weights matrices ($\mathbf{W}_{*(xr)}$ and $\mathbf{W}_{c}$) are stored in BRAM 2, 3. BRAM 4 stores bias vectors $\mathbf{b}$ and the projection matrix ${\mathbf{W}}_{ym}$ is stored in BRAM 5. Of course these weight matrices are stored with compression in the block-circulant framework.

Based on data dependency of the LSTM model, we propose to adopt \emph{multi-stage coarse-grained pipelining} (abbreviated as CGPipe) techniques, to achieve maximum performance under the resource constraints. The first CGPipe stage is responsible for multiplication of weights matrices (i.e.,$\mathbf{W}_{*(xr)}$) and input vectors $[\mathbf{x}_t^T, \mathbf{y}_{t-1}^T]^T$. The second CGPipe stage is in charge of non-matrix vector multiplications such as diagonal matrix-vector multiplication, bias addition, and activation functions. The third CGPipe stage processes the matrix-vector multiplication for projection matrix $\mathbf{W}_{ym}$ and projected output $\mathbf{y}_{t}$. A double buffer is inserted among each CGPipe stage to shorten the idle time. Fine-grained pipelining (abbreviated as FGPipe) methodology is utilized to schedule the associated sub-operations for each CGPipe stage. In our designs, double buffers are only used between each pair of concatenated coarse-grained pipelining stages and only 3 coarse-grained stages are used. Double buffers are not used for weights. Because the inputs/intermediate results of LSTM/GRU do not have high dimension (with dimension of 1,024, as example), the double buffers only account for a very small portion of BRAM resource.
 
The intermediate results ($ \mathbf{c}_t$ and $\mathbf{m}_t$) are initialized to zero. To explain the mechanism of the architecture, we take the computation of forget gate $\mathbf{f}_{t}$ as a demonstration. As shown in Fig.~\ref{fig:lstm_arch}, input feature vectors $[\mathbf{x}_t^T, \mathbf{y}_{t-1}^T]^T$ fetched from BRAM 1 and weight matrices $\mathbf{W}_{f(xr)}$ fetched from BRAM 2
% ($[\mathbf{W}_{cr}, \mathbf{W}_{cr}]$)
are prepared for PEs for the purpose of calculating $\mathbf{W}_{fx} \mathbf{x}_t$ and $\mathbf{W}_{fr} \mathbf{y}_{t-1}$ in CGPipe stage 1. $\mathbf{W}_{fc} \mathbf{c}_{t-1}$ is generated by point-wise multiplication (a group of multipliers) in the first phase of CGPipe stage 2. Adder trees accumulate $\mathbf{W}_{fx} \mathbf{x}_t$,  $\mathbf{W}_{fr} \mathbf{y}_{t-1}$, $\mathbf{W}_{fc} \mathbf{c}_{t-1}$, and bias $\mathbf{b}_{f}$ in the second phase of CGPipe stage 2. After passing the intermediate data through the activation function $\sigma$, E-RNN produces the result $\mathbf{f}_{t}$. The computations of other gates are implemented similarly. In the third phase of CGPipe stage 2, the computed gate outputs ($\mathbf{i}_{t}$, $\mathbf{g}_{t}$, and $\mathbf{f}_{t}$) are then fed into the multiplier-adder block. By multiplying $\mathbf{o}_{t}$ with the intermediate result from $tanh$ activation, E-RNN produces the projected output $\mathbf{m}_{t}$. Output $\mathbf{y}_{t}$ will be written back to BRAM 1 and replace $\mathbf{y}_{t}$ for the next recurrent process ($\mathbf{y}_{t-1} \leftarrow \mathbf{y}_{t}$) after CGPipe stage 3.  

\subsubsection{CU Implementation of GRU}

  \begin{figure}[t]
    \centering
    \includegraphics[width=0.65\columnwidth]{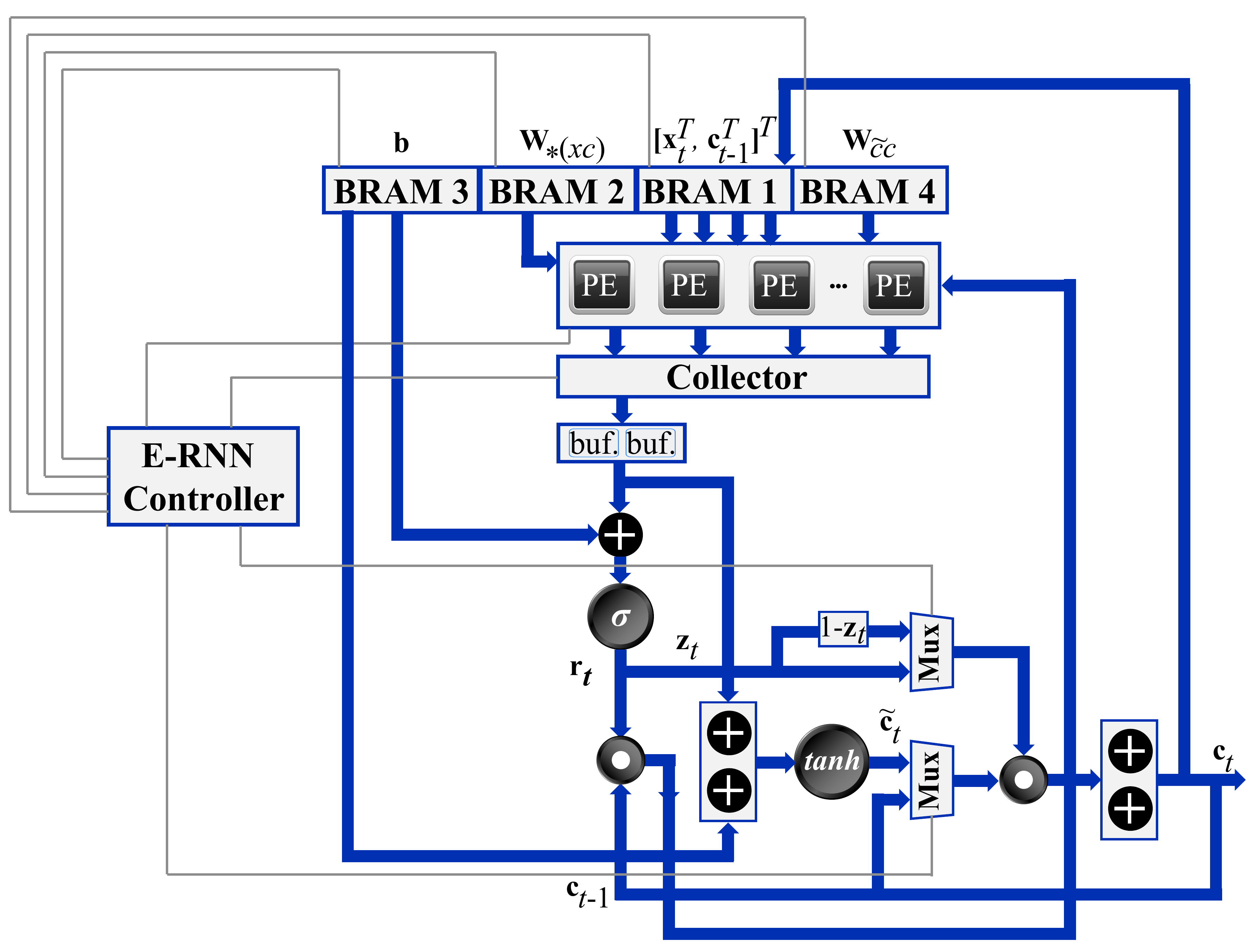} 
    \vskip -0.5em
    \caption{A compute unit (CU) with multiple processing elements (PEs) of GRU.}
    \label{fig:gru_arch}
 \end{figure}

The CU of GRU model described in Eqn. (\ref{eqn:grumodel}) can also be implemented using above design. The proposed architecture for GRU is shown in Fig.~\ref{fig:gru_arch}, 
which contains multiple PEs, double buffer, sigmoid/tanh, adder tree, and element-wise multiplier. GRU architecture has four BRAM blocks, in which input feature vectors $[\mathbf{x}_t^T, \mathbf{c}_{t-1}^T]^T$ are stored in BRAM 1. 
Weight matrix $\mathbf{W}_{\ast(xc)}$ is stored in BRAM 2.
Bias values (including $\mathbf{b}_z$, $\mathbf{b}_r$, and $\mathbf{b}_{\tilde{c}}$) are stored in BRAM 3, and weight matrix $\mathbf{W}_{\tilde{c}x}$ is stored in BRAM 4.   

Multi-stage CGPipe techniques are utilized based on data dependency of the GRU model, to separate the timing and resource-consuming matrix-vector operations. In GRU, the first CGPipe stage takes charge of multiplication of $\mathbf{W}_{*(xc)}[\mathbf{x}_t^T, \mathbf{c}_{t-1}^T]^T$. The second CGPipe stage computes the multiplication of $\mathbf{W}_{\tilde{c}c}(\mathbf{r}_t \odot  \mathbf{c}_{t-1})$ ($\mathbf{r}_t$ calculated in the first CGPipe stage) and $\mathbf{W}_{\tilde{c}x}\mathbf{x}_t$. The third CGPipe stage is responsible for the point-wise multiplication, activation functions, and summation operations. In the proposed GRU architecture, CGPipe stage 1 and CGPipe stage 2 can be implemented using the same hardware resource of FPGA with TDM method.
% We also insert double buffer between CGPipe stages.  FGPipe techniques are used to schedule the associated sub-operations for each CGPipestage.

% Here we explain the data transfer process in the GRU architecture. $\mathbf{c}_{t-1}$ is initialized to zero.  $\mathbf{z}_{t}$ is computed as follows: E-RNN fetches input feature vectors $[\mathbf{x}_t^T, \mathbf{c}_{t-1}^T]^T$ from BRAM 1 and weight matrix $\mathbf{W}_{*(xc)}$ from BRAM 2. The fetched data are then transferred to PEs for matrix-vector multiplication. The intermediate results $\mathbf{W}_{*(xc)}[\mathbf{x}_t^T, \mathbf{c}_{t-1}^T]^T$ and bias $\mathbf{b}_{z}$ are then transferred to adder tree. After passing the intermediate summation results to the activation function $\sigma$, E-RNN will generate $\mathbf{z}_{t}$. The computation of $\mathbf{r}_{t}$ has the same process. The computation of $\mathbf{W}_{\tilde{c}c}(\mathbf{r}_t \odot  \mathbf{c}_{t-1})$ and $\mathbf{W}_{\tilde{c}x}\mathbf{x}_t$ can be implemented using the same data path  as $\mathbf{W}_{*(xc)}[\mathbf{x}_t^T, \mathbf{c}_{t-1}^T]^T$. These two intermediate results and bias values $\mathbf{b}_{\tilde{c}}$ are then transferred to adder tree and activation function $tanh$ sequentially. Then E-RNN generates $\mathbf{\tilde{c}}_{t}$. The point-wise multiplication and summation will output the final result $\mathbf{c}_{t}$ of E-RNN, which will be written back to BRAM 1 and replace $\mathbf{c}_{t-1}$ for the next loop ($\mathbf{c}_{t-1} \leftarrow \mathbf{c}_{t}$). 

\subsection{Input and Weight Quantization.}

% It is well known that short fixed-point arithmetic representations 
To achieve significant reduction in memory bandwidth and footprint compared to long floating-point numbers, in E-RNN, we adopt fixed-point arithmetic units instead of floating-point units. However, shorter bit width may result in dramatic accuracy degradation. Therefore, we need to carefully select the total number of the bits for fixed-point number representation, such that the LSTM/GRU model can be compressed with small accuracy degradation. In the inputs and weights quantization phase, we first analyze the numerical range of inputs and trained weights in LSTM/GRU, and then initialize the integer and fractional part. 
{The quantization levels are determined by the (i) range of FFT results, and (ii) the predefined number of quantization levels. Each layer has an additional static scaling factor, which will not increase hardware implementation complexity because the scaling factor will be stored along with the FFT results after quantization.}

The accuracy degradation from input/weight quantization is very small (i.e., \textless0.1\%) and will not affect the accuracy of the design. 12-bit weight quantization is in general a safe design (it is also used in ESE). 

\section{Evaluation and Results}

\begin{table*}[t]
% \scriptsize

  \centering
   \caption{Detailed comparisons for different (LSTM and GRU) RNN designs on FPGAs (ours, ESE, and C-LSTM).} %\vspace{-0.2cm}
  \vskip -0.8em
  \label{table:Exp-Result}
  
  \resizebox{1.95\columnwidth}{!}{
  \begin{threeparttable}
  \begin{tabular}{|c|c|c|c|c|c|c|c|c|c|c|}
    \hline
    & \textbf{ESE~\cite{han2017ese}} & \makecell{\textbf{C-LSTM FFT8 \cite{Wang2018clstm}}\\(Block size: 8)} & \multicolumn{2}{c|}{\makecell{\textbf{E-RNN FFT8}\\(Block size: 8)}} & \multicolumn{2}{c|}{\makecell{\textbf{E-RNN FFT16}\\(Block size: 16)}} &  \multicolumn{2}{c|}{\makecell{\textbf{E-RNN FFT8}\\(Block size: 8)}} & \multicolumn{2}{c|}{\makecell{\textbf{E-RNN FFT16}\\(Block size: 16)}} \\
   \hline
   \textbf{\makecell{RNN Cell}} & \multicolumn{6}{c|} {LSTM-1024 w/ projection-512~\cite{sak2014long, han2017ese}} & \multicolumn{4}{c|} {GRU-1024}\\
   \hline

   \textbf{\makecell{Matrix Size\\(\#Params of top layer)}} & 0.73M & \multicolumn{3}{c|}{0.41M}& \multicolumn{2}{c|}{0.20M} & \multicolumn{2}{c|}{0.45M} & \multicolumn{2}{c|}{0.23M} \\
   \hline
  
   \textbf{Quantization} & 12bit fixed & 16bit fixed & \multicolumn{8}{c|}{12bit fixed}\\ 
   \hline

   \textbf{\makecell{Matrix \\ Compression Ratio}} & 4.5 : 1\tnote{a} & \multicolumn{3}{c|}{7.9 : 1\tnote{c}} & \multicolumn{2}{c|}{15.9 : 1} &  \multicolumn{2}{c|}{8.0 : 1} & \multicolumn{2}{c|}{15.9 : 1}\\
   \hline

   \textbf{Platform} & KU060 & 7V3 & KU060 & 7V3 & KU060 & 7V3 & KU060 & 7V3 & KU060 & 7V3 \\
   \hline

   \textbf{DSP (\%)}  & 54.5 & 74.3 & 95.4 & 85.6 & 96.4 & 79.6 &  79.0  & 62.1 & 79.5 & 64.3  \\
   \hline

   \textbf{BRAM (\%)} & 87.7 & 65.7 & 88.1 & 78.5 & 90.3 & 65.2 & 90.8 & 88.2 & 81.2 & 79.5  \\
   \hline

   \textbf{LUT (\%)}  & 88.6 & 58.7 & 77.6 & 74.0 & 76.5 & 59.4  & 81.2 & 78.8 & 72.5 & 67.4  \\
   \hline

   \textbf{FF (\%)}   & 68.3 & 46.5 & 61.2 & 52.3 & 65.1 & 55.3  & 72.4 & 73.2 & 65.2 & 60.3  \\
   \hline

   \textbf{Frequency (MHz)} & \multicolumn{10}{c|}{200}\\
   \hline

%   \textbf{PER (\%)} & 20.7 & 24.57 & 24.57 & 25.48 & 25.48 & xx & xx & xx & xx\\
%   \hline

   \textbf{\makecell{PER Degradation}} & 0.30\% & 0.32\% & \multicolumn{2}{c|}{0.14\%} & \multicolumn{2}{c|}{0.31\%} & \multicolumn{2}{c|}{0.18\%} & \multicolumn{2}{c|}{0.33\%}\\
   \hline

   \textbf{Latency ($\mu$s)} & 57.0 & 16.7 & 13.7 & 12.9 & 7.4 & 8.3 & 10.5 & 10.5 & 6.7 & 6.5  \\
   \hline

   \textbf{\makecell{Frames per \\Second (FPS)}} & 17,544\tnote{b} & 179,687 & 231,514 & 240,389 & 429,327 & 382,510 & 284,540 & 284,463 & 445,167 & 464,582\\
   \hline
%   \textbf{Dense FPS} &  \multicolumn{5}{c|}{CPU: 166 \qquad GPU: 4161} &  \multicolumn{4}{c|}{CPU: \qquad GPU:}\\
%   \hline
   \textbf{Power (W)} & 41 & 22 & - & 24 & - & 25 & - & 22 & - & 29\\
   \hline

   \textbf{\makecell{Energy Efficiency \\(FPS/W)}} & 428 & 8,168 & - & 10,016 & - & 15,300 & - & 12,930 & - & 16,020 \\
   \hline
  \end{tabular}%\vspace{-0.6cm}
  \begin{tablenotes}
            \item[a] This estimation considers both weights and indices (there is at least one index per weight after compression in ESE). However, this is a pessimistic estimation for ESE because indices can use fewer bits for representation than weights.\\
            \item[b] We use ESE's theoretical computation time to calculate FPS, the real computation time is larger than theoretical one which leads to smaller FPS.\\
            \item[c] We measure the compression ratio by the number of parameters in matrices. As the network architectures are identical in C-LSTM and E-RNN, their matrix compression ratios are the same.
 \end{tablenotes}
  \end{threeparttable}
}
\vspace{-1.5em}
\end{table*}

\subsection{Evaluation Platform and Exploration}

 \subsubsection{Experimental Platform}
We use two FPGA platforms for evaluating the proposed E-RNN framework for LSTM and GRU RNNs: Alpha Data's ADM-PCIE-7V3 and Xilinx KU060. The ADM-PCIE-7V3 board, comprising a Xilinx Virtex-7 (690t) FPGA and a 16GB DDR3 memory, is connected to the host machine through PCIE Gen3 $\times$ 8 I/O Interface. Xilinx KU 060 is a Kintex UltraScale serial FPGA with two 4GB DDR3 memory. The host machine adopted in our experiments is a server configured with multiple Intel Core i7-4790 processors. The detailed comparison of on-chip resources of the two FPGA platforms is presented in Table~\ref{tbl:platform}. We use Xilinx SDX 2017.1 as the commercial high-level synthesis backend to synthesize the high-level (C/C++) based RNN designs on the selected FPGAs. The E-RNN framework of FPGA implementation of (LSTM and GRU) RNNs are operating at 200MHz on both platforms, which is configured to be the same as the prior works ESE \cite{han2017ese} and C-LSTM~\cite{Wang2018clstm} for fair comparisons.

\begin{table}[!ht]
    \centering
    \caption{Comparison of two selected FPGA platforms}\label{tbl:platform}
    \vskip -0.8em
    \resizebox{0.9\columnwidth}{!}{
        \begin{tabular}{|c|c|c|c|c|c|}
            \hline
            FPGA Platform & DSP & BRAM & LUT & FF & Process \\ 
            \hline
            ADM-PCIE-7V3 & 3,600 & 1,470 & 859,200 & 429,600 & 28nm\\
            \hline 
            XCKU060 & 2,760 & 1,080 & 331,680 & 663,360 & 20nm\\
            \hline

        \end{tabular}
    }
\end{table}

\subsubsection{High-Level Synthesis (HLS) Exploration}

\begin{figure}[b]
    \centering
    \includegraphics[width=0.95\columnwidth]{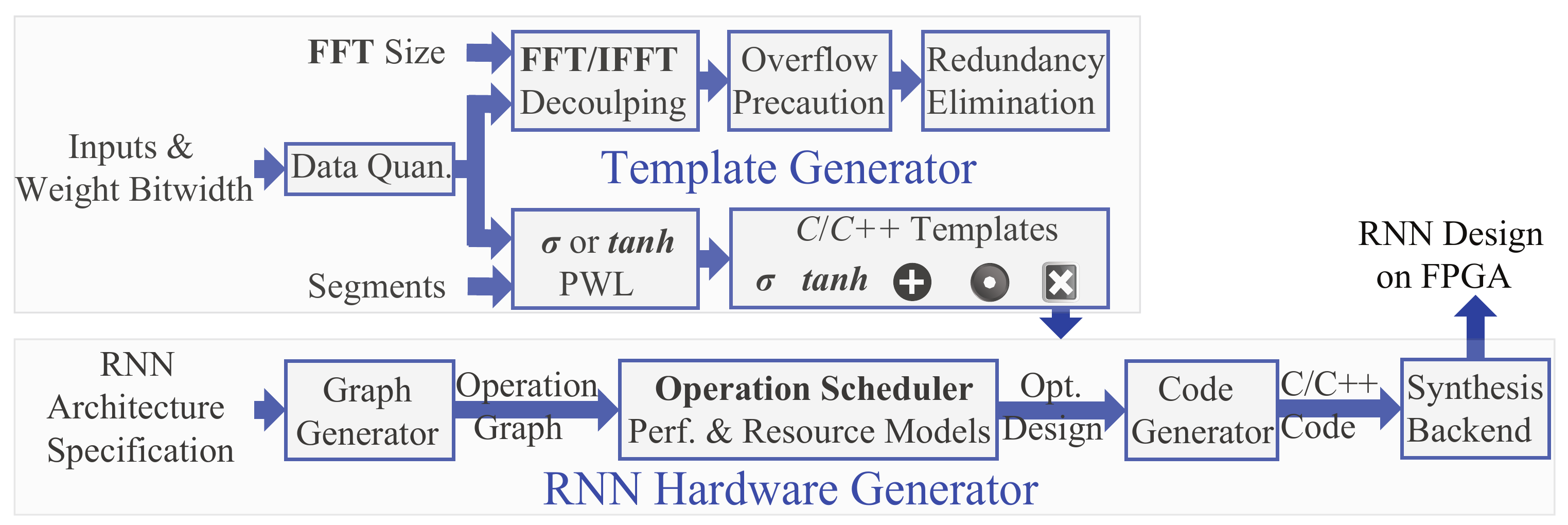} 
    \vskip -0.5em
    \caption{Overview of high level synthesis framework.}
    \label{fig:HLSframework}
 \end{figure}
 
We have developed an HLS framework for automatically converting high-level descriptions of RNNs into FPGA implementations, with the framework overview shown in Fig. \ref{fig:HLSframework}. This is a template-based framework for design automation of RNN implementations, based on the above described optimizations. 
The HLS framework consists of two parts which are the \emph{primitive operation templates generator} and the \emph{RNN hardware design generator}.
% Since there are a limited number of primitive operations in RNN models, we generate C/C++ based templates for these operations, with tunable parameters (e.g., bit-length, FFT module size, block size, etc.). Then the RNN hardware design generator is fed with the RNN architecture specification, which are the block-circulant matrix-based RNN models.
% A directed acyclic data dependency and operation graph is generated representing the computation flow in RNNs.
% The graph is then scheduled into multi-stage pipeline to maximize the performance under certain resource constraints with the help of accurate performance and resource models.
% The scheduling result is then fed to code generator and synthesis backend to output the final FPGA implementations. 
More details are provided as follows:

\textbf{\emph{Template Generator:}}
We develop the C/C++ based template for each of the primitive operations in RNNs, e.g., $tanh$, sigmoid $\sigma$, point-wise vector addition, point-wise multiplication, and ``FFT$\rightarrow$element-wise multiplication $\rightarrow$IFFT" procedure.

\textbf{\emph{Graph Generator:}}
%A directed acyclic data dependency and operation graph (like the one in Fig. \ref{fig:Data_dependency}(a)) is generated representing the computation flow in RNNs.
In order to extract the complicated interactions among primitive operations in an RNN model, we design a graph generator that produces a directed acyclic data dependency and operation graph unrolling the computations in RNNs.
We deliberately remove the feedback edges of $\mathbf{c}_{t}$ and $\mathbf{y}_{t}$, which are taken care of by the \emph{double-buffer mechanism}, and therefore do not harm the correctness and efficiency of the RNN.

\textbf{\emph{Operation Scheduler:}}
The computational complexities of the primitive operations in RNN exhibit a highly skewed distribution.
For example, the complexity of matrix-vector multiplication $[\mathbf{W}_{\ast x}\ \ \mathbf{W}_{\ast r}][\mathbf{x}_{t}^T, \mathbf{y}_{t-1}^T]^T$ is 128$\times$ as that of point-wise multiplication $\mathbf{W}_{ic} \odot \mathbf{c}_{t-1}$.
% If we want to pipeline these two operations, we must either boost the parallelism of former operation or make the latter operation wait (idle) for the former operation.
% However, on-chip resources of FPGA generally cannot support sufficient parallelism and the idle operations will make the design inefficient.
Therefore, we develop an automatic operation scheduler to generate a pipeline scheme given the data dependency and operation graph from the graph generator.
The objective is to maximize throughput under hardware resource constraints. 
% Details will be provided in a technical report due to space limitation.

\textbf{\emph{Code Generator and Synthesis Backend:}}
The code generator takes the operation scheduling result as input and generates the final C/C++ code automatically by integrating the involved primitive operations. 
The generated C/C++ code for RNN is then fed to an off-the-shelf commercial synthesis backend to generate the FPGA implementation.

\subsection{Experimental Results and Discussions}

We evaluate the performance on both FPGA platforms for LSTM and GRU RNNs using the same TIMIT dataset, which is the same dataset utilized in the prior works ESE and C-LSTM. 
The latencies of E-RNN framework implementation are measured by the total number of clock cycles ($N_{CC}$) multiplied by the clock period $T$ (5 $n$s) from the Xilinx SDx tools, and power/energy consumptions are from actual power measurements. 
For platform KU060, since we do not have the physical platform for power measurement, we leave the power and energy efficiency values to be blank in Table III. % We aim to demonstrate that E-RNN outperforms the previous FPGA-based LSTM accelerators ESE~\cite{han2017ese} and C-LSTM~\cite{Wang2018clstm} in terms of both performance and energy efficiency under the \uline{same accuracy degradation}.

As shown in Table~\ref{table:Exp-Result} with detailed comparison results, we explore on both LSTM and GRU, with two different block sizes 8 and 16, on both selected FPGA platforms. The bit length is optimized to be 12 bits, which is validated to result in no additional accuracy degradation due to quantization. We use the same baseline LSTM model with ESE/C-LSTM. 
(i) We present a comparison between E-RNN with block size 8 and ESE, in which case the compression ratio will be similar. The comparison aims to demonstrate the lower accuracy degradation and higher performance achieved by E-RNN; (ii) we present a comparison between E-RNN with block size 16 and ESE, in which case the accuracy degradation will be similar. The comparison aims to demonstrate that E-RNN achieves better performance and energy efficiency under the same accuracy degradation; (iii) we compare the performance and energy efficiency between E-RNN and C-LSTM using the same block size (both are based on the block-circulant matrix-based framework), to illustrate the effectiveness of the design optimization framework; (iv) we provide the results of E-RNN based on GRU model, for further enhancement on performance and energy efficiency.  

\subsubsection{Comparison with ESE}

When the block size is 8, the compression ratio of E-RNN is similar compared with ESE.
The comparison results, as shown in the first and third columns of Table~\ref{table:Exp-Result}, are both on the KU060 FPGA platform. We could observe that the E-RNN achieves lower accuracy degradation compared with ESE (0.14\% vs. 0.30\%), demonstrating the effectiveness of the block-circulant framework in terms of accuracy. We can also observe that E-RNN achieves 13.2$\times$ performance improvement, with an energy efficiency improvement of 23.4$\times$ using actual measurement results on the ADM-PCIE-7V3 board. It is necessary to note that as shown in Table~\ref{tbl:platform}, the manufacturing process of XCKU060 FPGA is 20nm while the process of Virtex-7 is 28nm, which means the energy efficiency gain reported here is even conservative.

Although the compression ratios are similar, the significant efficiency improvement is because of the following two reasons. First, the block-circulant framework results in a regular network structure, and therefore a significantly higher degree of parallelism. As an illustrative example, we can implement in parallel 16 FFTs, each with 16 inputs, in parallel in FPGA. In contrast, it will be especially difficult for ESE to operate in parallel $16\times 16=256$ inputs when the network is stored in the irregular structure (one weight indexing another). The second reason is the efficient implementations of tanh and sigmoid activation functions. Our piecewise linear approximation method can support activation implementation only using on-chip resources. In contrast, the ESE implements activations in look-up tables, and therefore requires off-chip DDR storage if enough parallelism is required (although it is possible to store all weight parameters of ESE on-chip). The latter reason accounts for more than 2$\times$ energy efficiency gain and the majority is attributed to the regularity benefit. As a side evidence, the LUT and FF utilizations of E-RNN are lower than ESE, which shows that E-RNN has less boolean and numeric nodes due to the regularity.

With block size 16, the accuracy degradation of E-RNN (using LSTM model) is similar as ESE. As shown in the first and fifth column of Table~\ref{table:Exp-Result}, the E-RNN achieves 24.47 $\times$ performance improvement, with a energy efficiency improvement of 35.75 $\times$ using ADM-7V3 platform compared with ESE. The results are at least 50\% higher than results of E-RNN with block size 8. 

% Moreover, E-RNN can achieve significant speedup and energy efficiency gain compared with CPU and GPU implementations. As baselines of ESE, the FPS values of CPU and GPU implementations are 166 and 4,161, respectively, for the baseline LSTM model without compression. In general, we observe that E-RNN can achieve real-time performance, overcome the challenges in RNN hardware implementations and the irregularity limitation of the weight pruning method.

\subsubsection{Comparison with C-LSTM}
We applied ADMM to well trained RNN models to train the block circulant matrices. As ADMM does not hurt the original model performance theoretically, but only convert the matrices to block circulant format, the accuracy degradation is smaller than C-LSTM. As a result, E-RNN achieves lower PER degradation than C-LSTM when given the same block size (0.14\% vs. 0.32\% with block size of 8).
We compare the performance and energy efficiency between E-RNN and C-LSTM using the same block size 8 (both are based on the block-circulant matrix-based framework). We can observe that E-RNN achieves 1.33$\times$ performance improvement with a block size of 8, with an energy efficiency improvement of 1.22$\times$ using the same ADM-PCIE-7V3 board. The similar observation is also obtained from comparison using block size of 16: E-RNN (using LSTM) achieves 1.32$\times$ performance and 1.06$\times$ energy efficiency improvement compared with C-LSTM.
These improvements are attributed to the design optimization framework, including hardware system design, PE optimization, and quantization. 
% (from 16 bit in C-LSTM to 12 bit in this work)

Among the three, the first two components are more effective compared to quantization: reducing from 16 bit to 12 bit only accounts for less than 10\% performance improvement. Compared to C-LSTM, E-RNN has a systematic architecture including PE and CU for both LSTM and GRU. 
In addition, the optimization target of E-RNN is in the bottom level, i.e., PE level.
% The adopted optimization methods include PE size/number optimization and reducing storage on PE intermediate results, etc. 
% Meanwhile, the PE optimization targets at the most resource-intensive arithmetic, i.e., matrix-vector multiplication. As a result, we can reduce a large number of multipliers and adders and achieve a higher degree of computation parallelism, which leads to the significant improvement of performance and energy efficiency in the FPGA implementation.
% The PE level optimization reduces a large number of multipliers and adders and achieves a higher degree of computation parallelism, which leads to the significant improvement of performance and energy efficiency in the FPGA implementation.
The seemingly counterintuitive observation is because the same number of DSP blocks are utilized in FPGA (on the other hand, BRAM does not account for a large portion of energy consumption in FPGA).

\subsubsection{Experimental Results on GRU}

As shown in the right four columns of Table~\ref{table:Exp-Result}, compared with ESE, C-LSTM, and E-RNN with LSTM, we can observe that the E-RNN with GRU model achieves 26.48$\times$, 2.59$\times$, and 1.21$\times$ performance improvement under the same accuracy degradation, respectively.
For the perspective of energy efficiency, the E-RNN with GRU model can achieve 37.4$\times$, 2.0$\times$, and 1.05$\times$ improvement, respectively. Experimental results show that the design optimization framework E-RNN with GRU model can have the best performance and energy efficiency. We verify that if the accuracy requirement can be satisfied, it is desirable to shift from LSTM to GRU because of less computation and storage.

\section{Conclusion}
In this paper, we use ADMM-based training for deriving block-circulant matrice-based RNN representation. We present the E-RNN framework for FPGA implementations of the ASR application. The overall goal is to improve performance/energy efficiency under accuracy requirement. 
We start from two design explorations providing guidance on block size and reducing RNN training trials. Based on the two observations, we decompose E-RNN in two phases: Phase I on determining RNN model to reduce computation and storage subject to accuracy requirement, and Phase II on hardware implementations given RNN model. We explore on both LSTM and GRU using the proposed E-RNN and we provide comprehensive comparisons with ESE and C-LSTM. Experimental results demonstrate the effectiveness of the proposed framework E-RNN compared with the prior works ESE and C-LSTM.

\section*{Acknowledgments}

This research is supported by the National Science Foundation grants NSF-CCF-1733701, NSF-CNS-1704662, NSF-CNS-1739748,  NSF-CCF-1657333, NSF-CCF-1717754, NSF-CNS-1717984, and NSF-CCF-1750656. We thank all the anonymous reviewers
for their feedback.

% use section* for acknowledgement
% \section*{Acknowledgment}

% The authors would like to thank...
% more thanks here

% trigger a \newpage just before the given reference
% number - used to balance the columns on the last page
% adjust value as needed - may need to be readjusted if
% the document is modified later
%\IEEEtriggeratref{8}
% The "triggered" command can be changed if desired:
%\IEEEtriggercmd{\enlargethispage{-5in}}

% references section

% can use a bibliography generated by BibTeX as a .bbl file
% BibTeX documentation can be easily obtained at:
% http://www.ctan.org/tex-archive/biblio/bibtex/contrib/doc/
% The IEEEtran BibTeX style support page is at:
% http://www.michaelshell.org/tex/ieeetran/bibtex/
%\bibliographystyle{IEEEtran}
% argument is your BibTeX string definitions and bibliography database(s)
%\bibliography{IEEEabrv,../bib/paper}
%
% <OR> manually copy in the resultant .bbl file
% set second argument of \begin to the number of references
% (used to reserve space for the reference number labels box)
% \begin{thebibliography}{1}

% \bibitem{IEEEhowto:kopka}
% H.~Kopka and P.~W. Daly, \emph{A Guide to \LaTeX}, 3rd~ed.\hskip 1em plus
%   0.5em minus 0.4em\relax Harlow, England: Addison-Wesley, 1999.

% \end{thebibliography}
\footnotesize
\bibliographystyle{ieeetr}
\bibliography{ref}

% that's all folks
\end{document}